%% file: ms.tex
\documentclass[9pt,technote]{IEEEtran} 

\IEEEoverridecommandlockouts                              

\usepackage{epsfig} 
\usepackage{amsmath} 
\usepackage{amssymb}  
\usepackage{dsfont}
\usepackage{graphicx}
\usepackage{booktabs,array}
\usepackage{rotating}
\usepackage{caption}
\usepackage{subcaption}
\usepackage{color}
\usepackage[table]{xcolor}
\usepackage{multirow}
\usepackage{tikz}
\usetikzlibrary{arrows,chains,matrix,positioning,scopes}
\usepackage{mathtools}

\makeatletter
\tikzset{join/.code=\tikzset{after node path={%
\ifx\tikzchainprevious\pgfutil@empty\else(\tikzchainprevious)%
edge[every join, bend right]#1(\tikzchaincurrent)\fi}}}
\makeatother
\tikzset{>=stealth',every on chain/.append style={join},
         every join/.style={->}}
\tikzstyle{labeled}=[execute at begin node=$\scriptstyle,
   execute at end node=$]

\newcommand\copyrighttext{%
\footnotesize Submitted to IEEE Transactions on Robotics, January 2018. \copyright 2018 IEEE. Personal use of this material is permitted. Permission from IEEE must be obtained for all other uses, in any current or future media, including reprinting/republishing this material for advertising or promotional purposes, creating new collective works, for resale or redistribution to servers or lists, or reuse of any copyrighted component of this work in other works.}
\newcommand\copyrightnotice{%
\begin{tikzpicture}[remember picture,overlay]
\node[anchor=south,yshift=7pt] at (current page.south) {\fbox{\parbox{\dimexpr\textwidth-\fboxsep-\fboxrule\relax}{\copyrighttext}}};
\end{tikzpicture}%
}

\title{\LARGE \bf Multiple Object Detection, Tracking \\
and Long-Term Dynamics Learning \\ in Large 3D Maps}

\author{\IEEEauthorblockN{Nils Bore, Patric Jensfelt and John Folkesson\\}
\IEEEauthorblockA{Robotics, Perception and Learning Lab\\
Royal Institute of Technology (KTH)\\
Stockholm, SE-100 44, Sweden\\
Email: \{nbore, patric, johnf\}@kth.se}}

\begin{document}

\maketitle
\copyrightnotice
\vspace{-15pt} 

\begin{abstract}

In this work, we present a method for tracking and learning
the dynamics of all objects in a large scale robot environment.
A mobile robot patrols the environment and
visits the different locations one by one.
Movable objects are discovered by change detection,
and tracked throughout the robot deployment.
For tracking, we extend the Rao-Blackwellized particle filter of \cite{bore2017tracking} with birth
and death processes, enabling the method to handle an arbitrary number of objects.
Target births and associations are sampled using \textit{Gibbs sampling}.
The parameters of the system are then learnt using the \textit{Expectation Maximization}
algorithm in an unsupervised fashion.
The system therefore enables learning of the dynamics
of one particular environment, and of its objects.
The algorithm is evaluated on data
collected autonomously by a mobile robot in
an office environment during a real-world deployment.
We show that the algorithm automatically identifies and tracks the
moving objects within 3D maps and infers plausible
dynamics models, significantly decreasing the modeling bias of our
previous work.
The proposed method represents an improvement over previous
methods for environment dynamics learning as it allows for learning of fine
grained processes.

\end{abstract}
\begin{IEEEkeywords}
Mobile robot, multi-target tracking, movable objects, dynamics learning.
\end{IEEEkeywords}

\IEEEpeerreviewmaketitle

\section{Introduction}

\input{introduction.tex}

\section{Related Work}

\input{related_work.tex}

\section{Method}

\input{method.tex}

\section{Experiments}

\input{experiments.tex}

\section{Results}

\input{results.tex}

\section{Discussion \& Future Work}

\input{discussion.tex}

\section{Conclusion}

\input{conclusion.tex}

\addtolength{\textheight}{-10.0cm}   




%




\section{Acknowledgements}

The work presented in this paper has been funded by the European
Union Seventh Framework Programme (FP7/2007-2013) under grant
agreement No 600623 (``STRANDS'').

\bibliography{toc_new}
\bibliographystyle{ieeetr}

\copyrightnotice

\end{document}

%% file: introduction.tex
Simple mobile robots are becoming prevalent, for example through
commercialization of lawn mower and vacuum cleaner robots.
In the research world, we are starting to see examples of
more complex service
robots that are interacting with humans in everyday environments such as 
museums \cite{burgard1999experiences}, airports \cite{triebel2016spencer},
offices \cite{veloso2012cobots} and care homes \cite{hawes2016strands}.
In these environments, there may be a high amount of dynamics
such as movable objects and humans or other autonomous agents.
Dynamics are often cited as a challenge to service robots
since it makes estimation harder and prevents accurate prediction \cite{alterovitz2016robot}.
The classical solution has been to integrate uncertainty into the
algorithms, and trade precise models for larger estimated uncertainty.
In this work, we investigate if a robot can learn more accurate dynamics models
using extended experiences from one environment. Ideally, if the robot
can learn models of the dynamics of cluttered environments, it
can itself tune the exact amount of modeling uncertainty it needs when
operating in these environments. In addition, prediction can be
greatly improved by learning typical environment motion.

\input{system}

As a practical motivation for this line of research, it is useful to consider the problem of
a general autonomous cleaning robot.
Such a system has been proposed as one of three
\textit{challenges of robotic manipulation} by Kemp et al. \cite{kemp2007challenges}.
In contrast to a vacuum cleaner, a general
robot should have the ability to pick up items, put them back in their places,
possibly fold up clothes, etc.
Therefore, it needs to observe the
clutter and reason about how to bring it into order.
The task requires an understanding
of the structure of a scene, with the objects in it.
Moreover, it requires the ability to discover new objects
and reason about the normal arrangements of these objects.
If a cleaning robot notices
any object in an unusual place, it should be able to put it back where it belongs.
Note that if a robot has this kind of a ability, it would go a long way towards solving
the issue of operation in dynamic environments.
In the following we will discuss our approach for modeling
such environments and how it might facilitate general cleaning.

In this work, we attempt to model and learn the movement of the individual components of the environment, that is, the objects.
Tracking of \textit{movable} or \textit{semi-static} objects
from a mobile robot is a hard problem. The main difficulty
lies in that the objects might be moved when the robot is not there.
Some of these movements are hard to capture within
any motion model, as a human might take potentially arbitrary
actions with the objects. But fortunately, most objects have
a clearly defined use case, often tied to some position
in the environment. Consider for example an office desk
with chair, monitor, lamp, computer etc. These objects
will rarely stray far from their usual placement, as it
is there that they are useful. Other objects such as mugs
might have purposes in multiple places in the environment
such as on the table or in the kitchen locker. The fact that
most objects mostly only move locally and in a few different
places allows us to build a dynamics model.

In our previous work \cite{bore2017tracking}, we described a system for 
maintaining position posteriors of a given set of objects in a
large scale robot environment.
Since the objects are typically located in different locations across the
environment, the model takes into account that the robot cannot
observe all the objects at once. The system also incorporates the fact
that objects may move to a different location when the robot is not
present to observe the movement. While the system described
in \cite{bore2017tracking} can track a fixed set of objects,
it cannot analyze the entire
robot environment, with a variable number of moving objects. In
addition, manual selection of model parameters means that it generally
needs to overestimate noise. In the current work, we improve on this
system by allowing an arbitrary number of objects as well as incorporating
online learning of model parameters.
By integrating the posterior probability of the object positions
over time, the robot can build a model of where the objects
are typically placed, and where they are not supposed to be.
The representation therefore provides the necessary information
for systems such as the general cleaning robot described above.
Moreover, the learning of model parameters allows the robot to
successively become more effective in modeling the environment,
thus improving its performance in general tasks.

Since the proposed method aims to improve a robot's environment
representation and capability to complete useful tasks, we implement
a complete robot pipeline, from sensor data to parameter learning.
Further, we evaluate performance on real-world data collected in
a controlled environment as well as in an actual robot deployment.
An overview of the system is presented in Figure \ref{fig:system}:
Our robot patrols the environment and gathers RGBD data in a
number of locations, mostly corresponding to rooms. Our method
identifies moving objects through change detection and temporal
reasoning. The tracking scheme presented in this work identifies
the number of objects and forms posteriors over the object positions
given the detections. When the tracker has assembled a history
of object positions, our learning scheme can iteratively learn the
parameters of the object movement, allowing us to refine our
posteriors.

Our contributions are the following:
\begin{itemize}
\item Building on \cite{bore2017tracking}, we demonstrate the feasibility of inferring the number of objects at the same time as tracking their jumps in a large environment
\item Learning of object instance dynamics, enabling better modeling of dynamic environments
\end{itemize}

%% file: system.tex
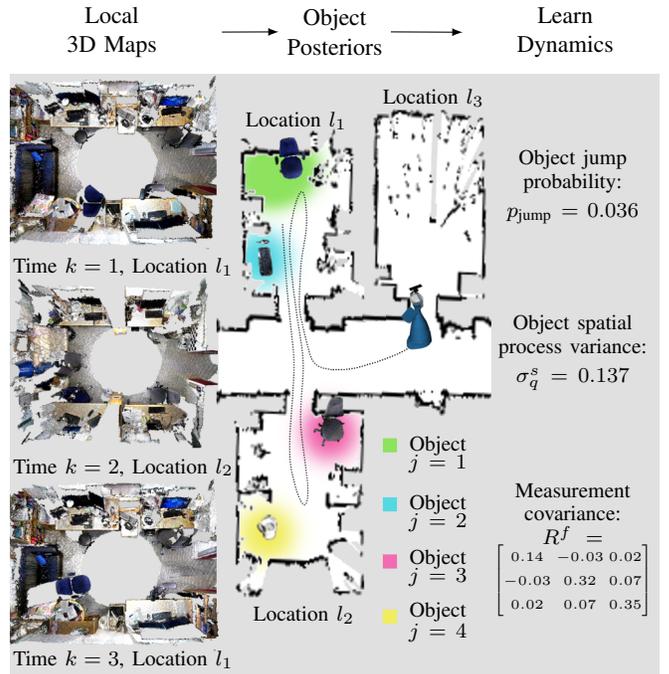
\begin{figure}
\begin{tikzpicture}
\node (img) {\includegraphics[width=0.97\linewidth]{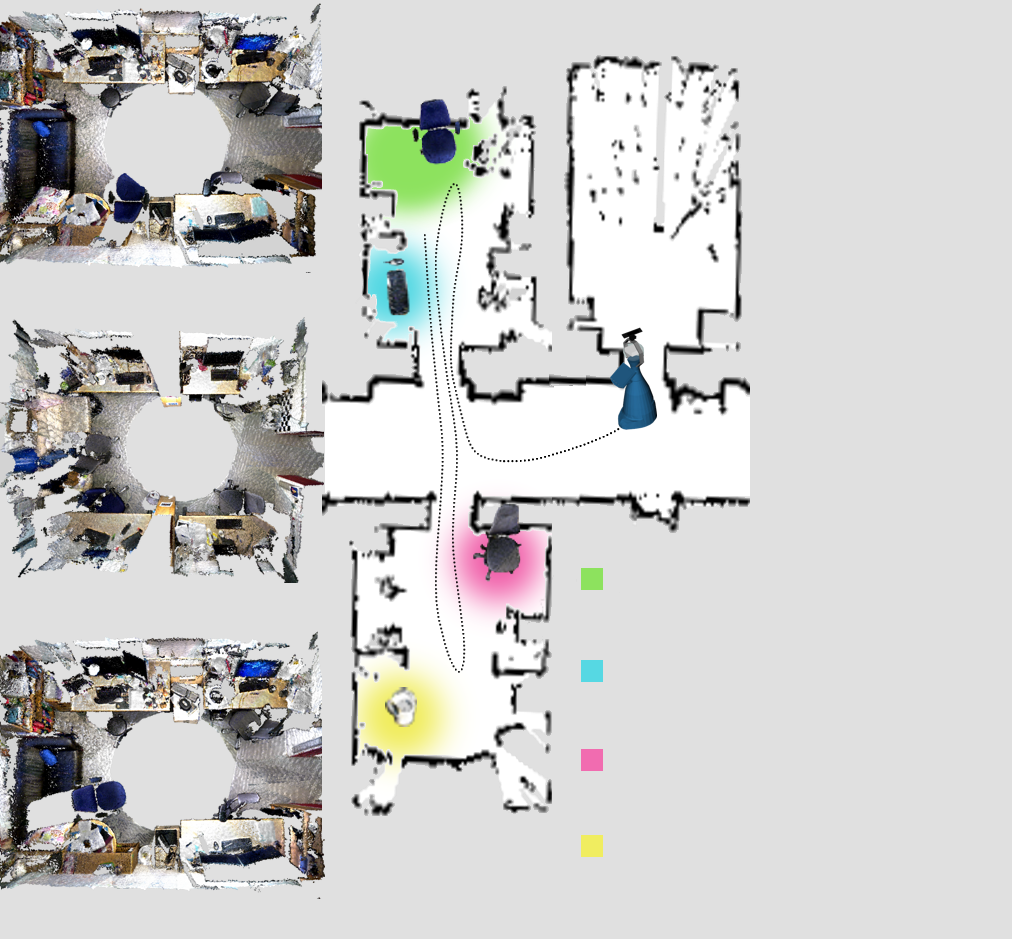}};
\node (obs) [above right,text width=2.7cm,align=center] at (img.north west){Local \\ 3D Maps};
\node (post) [above,align=center] at ([yshift=0.035cm]img.north){Object \\ Posteriors};
\node (learn) [above left,text width=2.5cm,align=center] at (img.north east){Learn \\ Dynamics};
\draw [->,>=latex] (obs.east) -- (post.west);
\draw [->,>=latex] (post.east) -- (learn.west);

\node (loc1) [below left,text width=1.4cm,align=center] at ([xshift=0.3cm, yshift=-0.5cm]img.north){\footnotesize  Location $l_1$};
\node (loc2) [below right,text width=1.4cm,align=center] at ([xshift=0.5cm, yshift=-0.2cm]img.north){\footnotesize  Location $l_3$};
\node (loc3) [above left,text width=1.4cm,align=center] at ([xshift=0.4cm, yshift=0.7cm]img.south){\footnotesize  Location $l_2$};

\node (math1) [below left,text width=2.3cm,align=center] at ([yshift=-1.0cm]img.north east){\footnotesize  Object jump \\ probability: \\ $p_\text{jump} = 0.036$};
\node (math2) [left,text width=2.3cm,align=center] at ([yshift=0.3cm]img.east){\footnotesize  Object spatial \\ process variance: \\ $\sigma_q^s = 0.137$};
\node (math3) [above left,text width=2.3cm,align=center] at ([yshift=0.8cm]img.south east){\footnotesize   Measurement \\ covariance: \\ $R^f =$ \\
\begingroup 
\setlength\arraycolsep{1pt}
\begin{tiny}
	 $\begin{bmatrix}
       0.14 & -0.03 & 0.02           \\
       -0.03 & 0.32 & 0.07 \\
       0.02 & 0.07 & 0.35
     \end{bmatrix}$
\end{tiny}
\endgroup
};

\node (obs3) [above right,text width=3.0cm,align=center] at ([yshift=0.1cm]img.south west){\footnotesize Time $k=3$, Location $l_1$};
\node (obs2) [above,text width=3.0cm,align=center] at ([yshift=2.1cm]obs3.north){\footnotesize Time $k=2$, Location $l_2$};
\node (obs1) [above,text width=3.0cm,align=center] at ([yshift=2.2cm]obs2.north){\footnotesize Time $k=1$, Location $l_1$};

\node (obj4) [right,text width=1.4cm,align=center] at ([xshift=4.5,yshift=-0.1cm]loc3.east){\footnotesize Object \\ \vspace{-0.12cm} $j=4$};
\node (obj3) [above,text width=1.4cm,align=center] at ([yshift=0.0cm]obj4.north){\footnotesize Object \\ \vspace{-0.15cm} $j=3$};
\node (obj2) [above,text width=1.4cm,align=center] at ([yshift=0.01cm]obj3.north){\footnotesize Object \\ \vspace{-0.15cm} $j=2$};
\node (obj1) [above,text width=1.4cm,align=center] at ([yshift=0.07cm]obj2.north){\footnotesize Object \\ \vspace{-0.15cm} $j=1$};
\end{tikzpicture}
\caption{Overview of the object tracking and dynamics learning
         system. At each time step $k$, the system collects a local 3D   
         map of one of the locations $l$ in its environment. From
         the observations we detect objects and form probabilities
         over their presence and positions. With many observations
         and experience of object movement, the system learns
         models of the environment dynamics. The improved models enable
         us to refine our environment representations.}
\label{fig:system}
\end{figure}

%% file: related_work.tex
In previous work \cite{bore2017tracking}, we investigated the problem of 
tracking a fixed number of objects in large robot environments.
Due to the environment scale, only some of
the objects are observed at one given time.
While the robot
is moving between different locations, the objects may move.
The robot thus has to reason about motion that it did not observe.
We addressed this problem with a principled \textit{Rao-Blackwellized
particle filter} (RBPF),
which samples data associations jointly using blocked Gibbs sampling \cite{geman1984gibbs}. 
The paper \cite{bore2017tracking} demonstrates the feasibility of tracking jumps of
many objects given that they are all present somewhere in the environment.
In the current work, we improve on the method by dropping
the closed world assumption, allowing tracking of a variable number
of objects. In addition, the proposed algorithm learns
the dynamics parameters in an unsupervised fashion.

In general, mobile robots need to perform localization and planning
to be able to move around in an environment. Historically,
the dynamics of most human environments have been perceived
as a challenge when performing these tasks \cite{alterovitz2016robot}.
Early approaches assumed a static environment, and discarded measurements
not agreeing with this model as noise. It has subsequently been
shown that static models works for many common environments. 
However, in highly dynamic scenarios,
this might lead to the system discarding vital information.
This has lead many researchers to pursue learning dynamics from
robot experience. For example, 
Kucner et al. \cite{kucner2013conditional} and Wang et al. 
\cite{wang2014modeling} both learn to predict occupancy maps
given previous map states. Meanwhile, Krajnik et al. 
\cite{krajnik2017fremen} have successfully applied their \textit{frequency
map enhancement} approach to several robotic problems. The method
allows them to estimate periodic dynamics in the maps which can also
be used to better model subsequent observations.
Another line of research aims to learn dynamics of the environment
in order to successfully manipulate it. Notable examples in this
area include Endres et al. \cite{endres2013learning}, who estimated
the dynamics of individual doors, and Scholz et al. 
\cite{scholz2015learning}, who produced several systems with
the aim of navigating an environment with obstacles blocking the
robot's way. Our work differs from that of previous methods in that
we learn dynamics of \textit{all objects} in the environment,
as opposed to for map cells 
\cite{kucner2013conditional}\cite{wang2014modeling}\cite{krajnik2017fremen}
or for individual objects by manipulation \cite{endres2013learning}\cite{scholz2015learning}.

Our approach to environment modeling is essentially
\textit{detection and tracking of multiple objects} (DATMO) \cite{wang2002simultaneous}, a term which
encompasses a broad area of work in robotic mapping. This area
can be further divided into two categories, one dealing with tracking of short-term dynamics
such as people and cars \cite{wang2007simultaneous}\cite{montemerlo2002conditional},
and the other with long-term dynamics, mostly \textit{semi-static}
objects in indoor environments.
Methods for long-term dynamics are directly related to our approach.
Most of them face the problem of only sporadically observing the objects,
and thus have to re-identify them with previous object estimates.
Early work was typically based on robot laser data. Schulz et al. 
\cite{schulz2001probabilistic} presented a system for tracking
objects moving a bit between observations,
and Wolf et al. proposed to incorporate tracking
directly into the occupancy map \cite{wolf2003towards}.
Gallagher et al. presented GATMO \cite{gallagher2009gatmo}, which
does not incorporate probabilistic motion models, but instead also
handles objects that move longer distances, for example between different rooms.
The authors describe the first system that is designed to keep track of
all discrete entities in an indoor environment, and it is close in
aim to what we are presenting here. In \cite{bore2017tracking},
we introduced a system for tracking a fixed number of objects, but which improved
on GATMO by formulating the problem probabilistically, enabling our
system to handle noisy observations. In this work, we further
improve on \cite{gallagher2009gatmo}\cite{bore2017tracking} by
also learning the dynamics of the environment as the robot gathers more data.
This is a conceptual development, as it allows our method
to adapt to individual environments as well as individual
objects within those environments.
Practically, the approach removes the need for
parameter tuning and improves robustness.

DATMO is also an instance of the \textit{multi-target tracking} (MTT) problem.
A popular and principled approach for estimating variable numbers of targets is
\textit{Random Finite Sets} (RFS) \cite{mahler2001multitarget}\cite{vo2006gaussian}.
In \cite{pasha2006closed}, Pasha, Vo et al. derived a closed form
solution to the jump Markov problem in the case of linear Gaussian dynamics,
and extended it further to non-linear systems in \cite{vo2006gaussian2}.
In \cite{punithakumar2008multiple}, a Sequential Monte Carlo PHD
filter with underlying models was derived,
allowing arbitrary underlying distributions and motion models.
Especially the jump Markov GMM-PHD \cite{pasha2006closed}\cite{vo2006gaussian}
filters are similar to our system. While they can not incorporate the
discontinuous motion of our environment, the SMC solution \cite{punithakumar2008multiple} can.
However, our learning scheme requires posteriors
over full target tracks for parameter estimation.
PHD filters are unsuited to this use case
as they instead maintain the \textit{intensity function} over all targets.
Another approach for probabilistic variable target tracking is to
incorporate birth and death processes that model the intensity with
which new targets appear and disappear from the tracked space
\cite{sarkka2007rao}\cite{kreucher2005multitarget}. These methods
employ classical Monte Carlo filters to recursively sample full posteriors
of the individual targets. Each particle in these schemes represents
a sample of the configuration of all targets. We build on these methods
in our application since they fulfil the stated requirements. One can
consider our approach similar to the \textit{Joint Multitarget Probability Density}
of Kreucher \cite{kreucher2005multitarget}, while incorporating \textit{Rao-Blackwellization}
of the continuous states as presented e.g. in \cite{sarkka2007rao}.

A few papers have dealt with modeling the robot's
belief of observing an object in a particular place, given past experience. 
Dayoub et al. \cite{dayoub2010toward} presented a system for inferring
the discrete locations of different objects. Their method is based on long-term
and short-term memories, with the recency of an observation
determining its relative importance. This leads the system to have
more confidence that an object in the same place as one of the later observations.
More recently, Toris et al. \cite{toris2017temporal} proposed
a model for object persistence. Essentially, they look at part
of our problem, namely estimating for how long an object observed
by a mobile robot can be expected to stay at its last location.
They propose modeling this period with an exponential distribution,
and also present a scheme for estimating its parameters from data.
Their results show that the proposed approach outperforms
simpler models on data that exhibits periodic movement.
In our system, we employ a simpler persistence model,
as it suffices to demonstrate the concept.
However, our filter would be able to accommodate the
model of \cite{toris2017temporal} by modifying the proposal distribution.

Though we put more emphasis on dynamics modeling,
the result of our inference is related to the concept of \textit{object instance discovery}.
Object instance discovery is the process of segmenting and clustering object observations
into classes of separate physical objects.
Finman et al. presented a full object discovery system in \cite{finman2013toward}.
Interestingly, their approach to modeling the features is similar to ours,
as they also assume the feature observations to include Gaussian noise.
The authors cluster the observations online and update the filter estimates with a Kalman filter.
Like most other works in this area, they do not take the position of the objects into account.
In \cite{ambrus2015unsupervised}, Ambrus et al. presented a system
that does object discovery within single locations, such as a room.
Their clustering algorithm incorporates temporal consistency,
ensuring that no objects observed at the same time are clustered.
Our approach associates objects between different observations, similar
to these systems. 
Where we differ is that we form probabilities over
current object positions, defined over the whole robot environment.
This means that we need to incorporate a
probabilistic dynamics model, and perform joint inference.
Unlike us, several object discovery systems also output some registered, fused
3D model of the aggregated observations \cite{choudhary2014slam}\cite{collet2015herbdisc}.

To summarize, we improve on our previous work \cite{bore2017tracking}
by extending it to be able to track a \textit{variable} number of targets.
In particular, this allows our method to take only a sequence of
3D maps as input, and return a list of tracked dynamic objects with
posteriors. Our learning method
improves on previous schemes for learning the
dynamics of an environment by having either wider applicability
\cite{endres2013learning}\cite{scholz2015learning}
or by learning more detailed models than previosly
\cite{kucner2013conditional}\cite{wang2014modeling}.
Learning also removes the need for tuning of some of the parameters
in \cite{bore2017tracking}.

%% file: method.tex
Our robot moves between $N_l$ different locations $l \in \mathcal{L}$ within a large environment.
At each time step $k \in 1, \dots, K$, there is a set of objects at each location, with no overlap between the different sets.
The robot can only observe the object set of one location
$l_k^y \in \mathcal{L}$ at each time $k$.
Each observation consists of a set of $M_k$ point measurements
$\mathbf{Y}_k = \{\mathbf{y}_{k, 1}, \dots, \mathbf{y}_{k, M_k} \}$,
corresponding to some or all of the objects at the location,
together with sporadic clutter measurements.
Each point measurement $\mathbf{y}_{k, m}$ includes a 2D position $\hat{\mathbf{y}}_{k, m}^s$ as 
well as a visual feature vector $\hat{\mathbf{y}}_{k, m}^f$.
We assume that the measurement position is observed with Gaussian noise with standard deviation $\sigma_r$.
Similarly, we assume that the feature measurement noise is Gaussian with covariance $\mathbf{R}^f$.
For our particular features, this is a reasonable assumption, 
as explained in Section \ref{sec:detections}.
A-priori, we do not know the movement direction, and instead assume
Gaussian process noise, with standard deviation $\sigma_q$, i.e. Brownian motion.
At each time $k$, each object might jump to a new discrete location $l \in \mathcal{L}$
with probability $p_\text{jump}$.
Our goal is to track the continuous positions of the objects,
and to infer how many movable objects are currently in the environment.
To track the movement,
we model the jumps in-between locations $l$
as a discrete Markov process, and use analytic models
for the continuous positions within the locations.
To handle a variable number of objects, we propose a model
of birth and death processes for each tracked object.

On top of tracking the objects, we want to learn the
parameters of the environment dynamics.
Within our framework, this amounts to learning the values of
$p_\text{jump}$, $\sigma_q$ and $\mathbf{R}^f$.
We employ the \textit{Expectation Maximization} (EM) algorithm
to learn these parameters, see Figure \ref{fig:em_algorithm}.
The expectation step consists of
the tracking algorithm described above being run on the full
sequence $1, \dots, K$.
Subsequently, the maximization step amounts to estimating the
parameters given the estimated positions and measurement associations.
The algorithm iteratively runs these two steps until convergence.
The full EM scheme is described in the text leading up to
Section \ref{sec:em_algorithm}.
In the following, we detail the tracking, and specifically the 
modeling of assignments of objects to locations and to point measurements.

\subsection{Jump Processes}
\label{sec:prior}
\input{prior_table}

In previous work \cite{bore2017tracking}, we proposed to decompose the
movement of the objects into
a local component and a global component. The idea is that the local
component models the common, small adjustments of objects in the environment
such as an office chair rotating or the computer mouse on a desk moving.
Meanwhile, objects move to completely new places far more seldom.
We can therefore model this global movement as a rare event.
In return, we can allow for a higher degree of
unpredictability in those rare motions.
At time $k$, the state of each object $j$ is therefore
separated into a global location $l_{j, k} \in \mathcal{L}$
and a local part $\mathbf{\hat{x}}_{j, k}$, as
well as an existence indicator $e_{j, k}$ that we will get
back to later:
$$\mathbf{x}_{j, k} = \left( \begin{bmatrix} \mathbf{\hat{x}}_{j, k}^s \\ \mathbf{\hat{x}}_{j, k}^f \end{bmatrix}, l_{j, k}, e_{j, k} \right).$$
The discrete part $l_{j, k} \in \mathcal{L}$ signifies the current
location of object $j$, while the continuous part $\mathbf{\hat{x}}_{j, k}$
consists of the 2D position $\mathbf{\hat{x}}_{j, k}^s$ and the object's feature
vector $\mathbf{\hat{x}}_{j, k}^f$.
We will begin by describing
the global movement and the birth and death processes of a single object,
both of which can be described by discrete Markov chains.
In Section \ref{sec:likelihood}, we discuss how to combine this model with the local process.

\input{img/model}

The movement between the locations $\mathcal{L}$ is
governed by a Markov process, where an object might
take the action $u_{j, k} = \text{jump}$
to one of the other locations at each time step with
probability $p_\text{jump}$ or stay,
$p(u_{j, k} = \text{no jump}) = 1 - p_\text{jump}$.
If it jumps, it may jump to any of the $N_l = | \mathcal{L} |$ locations uniformly,
$p(l_{j, k} = l | u_{j, k} = \text{jump}) = \frac{1}{N_l}$.

We take $c_{j, k}$ to mean which of the $M_k$ measurements
from time $k$ that target $j$ gave rise to, if any.
If the current measurement location $l_k^y$
coincides with the current target location $l_{j, k}$,
target $j$ produces a particular measurement $m \in 1, \dots, M_k$ with probability
$p(c_{j, k} = m | l_{j, k} = l_k^y) = \frac{1}{M_k} p_\text{meas}$.
It may also produce no measurement,
$p(c_{j, k} = \epsilon | l_{j, k} = l^y_k) = 1 - p_\text{meas}$.
The robot cannot observe an object if it is
not at its current location, $p(c_{j, k} = \epsilon | l_{j, k} \neq l^y_k) = 1$.

We can combine each of these simple probabilities
into the individual target transition priors
$p(u_{j, k}, l_{j, k}, c_{j, k} | l^y_k, l_{j, k-1}) = p(c_{j, k} | l^y_k, l_{j, k}) p(l_{j, k} | l_{j, k-1}, u_{j, k}) p(u_{j, k})$,
which is fully expanded as part of Table \ref{table:prior}.

\subsection{Birth and Death Processes}
\label{sec:birth}

In this paper, we extend the model from \cite{bore2017tracking}
to also include a varying number of targets.
Formally, we will therefore have a countable infinite number of of targets,
$j \in \mathbb{N}$.
For each target $j$, we extend the state space with an existence indicator
$e_{j, k} \in \{\text{alive}, \text{dead}, \text{unborn}\}$.
The variable $e_{j, k}$ tells us if target $j$ is anywhere
within the domain at time step $k$ (alive), if it has disappeared (dead) or
has never existed (unborn).
The setup is similar to several previous previous works \cite{sarkka2007rao}\cite{kreucher2005multitarget}.
Assuming the births and deaths are independent of the
location of the object or the last action,
the individual prior of one target then turns into
\begin{equation*}
\begin{gathered}
p(e_{j, k}, u_{j, k}, l_{j, k}, c_{j, k} | e_{j, k-1}, l_{j, k-1}) \\
= p(u_{j, k}, l_{j, k}, c_{j, k} | e_{j, k}, l_{j, k-1}) p(e_{j, k} | e_{j, k-1}).
\end{gathered}
\end{equation*}
This means that the target propagates as described whenever
$e_{j, k-1} = \text{alive}$. If the object dies, the other
properties are irrelevant. If an object is born,
it has to be associated with a measurement at the time of birth, see
Table \ref{table:prior}.
The birth and death process
is defined as follows:
\begin{equation*}
p(e_{j, k}=\text{alive} | e_{j, k-1}) = \left\{
  \begin{array}{ll}
    1-p_\text{death}, \text{ if } e_{j, k-1}=\text{alive}\\
    p_\text{birth}, \text{ if } e_{j, k-1}=\text{unborn} \\
    0, \text{ if } e_{j, k-1}=\text{dead} \\
  \end{array}.
\right. \\
\end{equation*}
and $p(e_{j, k}=\text{dead} | e_{j, k-1}) = 1 - p(e_{j, k}=\text{alive} | e_{j, k-1})$.
Each point measurement $m \in 1, \dots, M_k$ may give birth to a new target,
thus limiting the potential number of births to $M_k$ at each time step.
A priori, without any knowledge of past measurements,
each of the $M_k$ potential new targets will therefore be born with probability $p_\text{birth}$.
At this stage, we do not consider which specific targets $j \in \mathbb{N}$ are born at each time step.
While the identity $j$ does not matter in this formal modeling, it plays a role in our
inference scheme and is described in Section \ref{sec:clustering}.
Finally, each target that is estimated to be alive may die
at time step $k$ with the probability $p_\text{death}$.

\subsection{Particle Filter}
\label{sec:filter_overview}

The particle filter introduced in \cite{bore2017tracking}
decomposes the posterior over the discrete global state,
$c_k = \left\{ e_{j, k}, l_{j, k}, c_{j, k} \right\}_j$, and the continuous local
state, $\mathbf{\hat{X}}_k = \left\{ \mathbf{\hat{x}}_{j, k} \right\}_j$
of positions and features, into
$p(\mathbf{\hat{X}}_k, c_k  | \mathbf{\hat{X}}_{k-1}, c_{1:k-1}, \mathbf{Y}_{1:k}) = p(\mathbf{\hat{X}}_k | \mathbf{\hat{X}}_{k-1}, c_{1:k}, \mathbf{Y}_{1:k}) p(c_k | \mathbf{\hat{X}}_{k-1}, c_{1:k-1}, \mathbf{Y}_{1:k}).$
In the current work, we extend the resulting
Rao-Blackwellized particle filter to work with
a variable number of targets.
In each iteration of the filtering, we sample the
global state from $p(c_k | \mathbf{\hat{X}}_{k-1}, c_{1:k-1}, \mathbf{Y}_{1:k})$,
giving us discrete samples $e_{j, k}^i, l_{j, k}^i, c_{j, k}^i$
for each particle $i$.
The target birth and death indicators $e_{j, k}$ introduced
here are sampled analogously to the other discrete variables,
see Section \ref{sec:inference}.
Given the sampled existence and assignment samples,
the continuous state components
$\mathbf{\hat{x}}_{j, k}^s, \mathbf{\hat{x}}_{j, k}^f$
are updated from the local state using
$p(\mathbf{\hat{X}}_k | \mathbf{\hat{X}}_{k-1}, c_{1:k}, \mathbf{Y}_{1:k})$.
They can be tracked using classical independent Kalman filters,
since each particle knows which target gave rise to which measurements,
and the noise is normally distributed.
Each particle $i$ thus also has one Kalman filter for each sampled existing object, with the current estimates denoted by $\boldsymbol{\mu}_{j, k}^i, \boldsymbol{\Sigma}_{j, k}^i$.
In our treatment, we will not detail the local process
inference further, and refer to \cite{bore2017tracking} for details.

\subsection{Joint Likelihood and Proposal}
\label{sec:likelihood}

Let us turn to the problem of formulating the filter
recursion, in which we sample the local state component $c_k$.
In order to define the update, we first need a measurement
likelihood.
The likelihood $\mathcal{L}(\mathbf{Y}_k | c_k^i)$ changes depending on
the current sampled joint associations and indicators $c_k^i$ (and implicitly $\boldsymbol{\mu}_{j, k-1}^i, \boldsymbol{\Sigma}_{j, k-1}^i$).
In our model, no two objects $j, j'$ can be associated with the same
measurement, $\nexists m: c_{j, k} = c_{j', k} = m$.
Given the full set of assignments $c_k^i$ of the targets, we
can infer which measurements have no object assignment
and must be explained as spurious clutter.
The point likelihood of such a measurement $m$ is given by the density
of the uniform background clutter $\mathcal{L}(y_{k, m} | c_k^i) = \frac{1}{S^f A_k}$.
The parameters $S^f$ and $A_k$ denote the support of
the feature and spatial uniform clutter distributions respectively.
Meanwhile, if it has a measurement association $c_{j, k} = m$,
the point likelihood is simply the Kalman marginal likelihood of the estimates of
$\mathbf{\hat{x}}_{j, k-1}^s, \mathbf{\hat{x}}_{j, k-1}^f$,
$\mathcal{L}(y_{m, k} | c_{j, k}^i = m) = \mathcal{L}_{KM}(\mathbf{y}_{m, k}; \boldsymbol{\mu}_{j, k-1}^i, \boldsymbol{\Sigma}_{j, k-1}^i)$.
The joint likelihood of all measurements 
$\mathbf{Y}_k$ is given by
\begin{equation*}
\mathcal{L}(\mathbf{Y}_k | c_k) = \prod_{m=1}^{M_k}  \mathcal{L}(y_{k, m} | c_k).
\end{equation*}

Assuming we have a \textit{joint} association prior $p(c_k | c_{1:k-1}, \mathbf{Y}_{1:k-1})$ defined over all targets,
as opposed to the \textit{individual} priors $p(e_{j, k}, u_{j, k}, l_{j, k}, c_{j, k} | e_{j, k-1}, l_{j, k-1})$,
it can be combined with
the likelihood to produce a filter recursion
\begin{equation*}
\begin{gathered}
p(c_{1:k} | \mathbf{Y}_{1:k}) \propto p(\mathbf{Y}_{k} | c_{1:k}, \mathbf{Y}_{1:k-1}) p(c_{1:k} | \mathbf{Y}_{1:k-1}) \\
=  \mathcal{L}(\mathbf{Y}_k | c_k) p(c_k | c_{1:k-1}, \mathbf{Y}_{1:k-1}) p(c_{1:k-1} | \mathbf{Y}_{1:k-1}).
\end{gathered}
\label{eq:update}
\end{equation*}
Maintaining the full posterior is
infeasible, as the space over $c_{1:k}$ grows exponentially 
with time.
This motivates our approximation of the update using the Rao-Blackwellized particle filter.
We will show how to recursively
sample from the posterior $p(c_k | \mathbf{Y}_{1:k})$
given samples from $p(c_{k-1} | \mathbf{Y}_{1:k-1})$ using
a proposal
\begin{equation}
q(c_k) \propto \mathcal{L}(\mathbf{Y}_k | c_k) p(c_k | c_{1:k-1}, \mathbf{Y}_{1:k-1})
\label{eq:proposal}
\end{equation}
that is directly proportional to the posterior update, yielding
minimal variance among the particle weights.
This is important in our scenario, as the state space is high-dimensional,
which can easily lead to particle depletion.
We will demonstrate how to sample from $q$ using
the simple independent priors from Section \ref{sec:prior}
as opposed to the full joint prior
$p(c_k | c_{1:k-1}, \mathbf{Y}_{1:k-1})$.

\subsection{Particle Updates and Weights}
\label{sec:inference}

We use blocked Gibbs sampling \cite{geman1984gibbs}
to sample from the proposal distribution
$q(c_k)$ (see Equation \ref{eq:proposal}).
It is important to sample at least two associations at a time,
in order to allow the measurements to switch associations
during one Gibbs iteration. Otherwise, some measurements could
get locked to one target although equally likely associations exist.
In each step of the MCMC chain, we therefore sample pairwise
conditional priors of two associations $c_{j, k}, c_{j', k}$ at a time, conditioned on the other assignments,
denoted $c_k^{-j,j'} = \left\{ c_{j^*, k} | j^* \in \mathbb{N} \right\} \setminus \left\{ c_{j, k}, c_{j', k} \right\}$.
The pairwise priors are given by
\begin{equation*}
q(c_{j, k}, c_{j', k} | c_k^{-j,j'}) \propto \left\{
  \begin{array}{ll}
    0, \text{ if } c_{j, k} = c_{j', k} \neq \epsilon \\
    \begin{array}{rr} \mathcal{L}(\mathbf{Y}_k | c_k) p(c_{j, k} | c_k^{-j,j'}, c_{k-1}) \\ \times \; p(c_{j', k} | c_k^{-j,j'}, c_{k-1})\end{array}, \text{ otherw.} \\
  \end{array}
\right. \\
\end{equation*}
with $p(c_{j, k} | c_k^{-j,j'}, c_{k-1})$ being the individual priors
from Section \ref{sec:birth}, with one small difference:
The value $M_k$ in the prior needs to be modified
to reflect the number of unassigned measurements, as given by $c_k^{-j,j'}$.
Note that $\mathcal{L}(\mathbf{Y}_k | c_k)$ does not need to be computed in
full, as only the relative scale between the assignments matter.
The scale is given by a product of two likelihoods, which
can be either Kalman marginal likelihoods, or clutter
likelihoods, depending on the assignments $c_{j, k}, c_{j', k}$.
The potential target births can
be handled in the same way as the existing targets, as the
association priors are defined equivalently, see Table \ref{table:prior}.
With $N_k$ targets and $M_k$ potential births, this means that there
are $N_k+M_k$ possible individual marginals to sample at each time $k$.
In each iteration, we sample the target pairs $j, j'$ uniformly among
those possibilities.
If an unborn target $j$ is associated with a measurement at the
end of the MCMC chain, it changes its indicator to $e_{j, k} = \text{alive}$.
For each time step and particle, we perform 25 iterations of
burn-in before collecting samples from $q(c_k)$.

The proposal distribution $q(c_k)$ is defined as proportional to the product
of our prior and likelihood. As that product does not sum to one,
it needs to be normalized to define the proposal $q(c_k)$.
Further, we need to compute it, since the normalization
$Z_k^i = \sum_{c_k} \mathcal{L}(\mathbf{Y}_k | c_k) p(c_k | c_{1:k-1}^i, \mathbf{Y}_{1:k-1})$
of the update 
differs between the different particle trajectories $i$.
The normalization $Z_k^i$ therefore defines our weight update.
For each particle weight $w_k^i$, the update step
is given by $w_{k+1}^i = Z_k^i w_k^i$, followed
by a normalization that ensures that the weights sum to one.
Importantly, $Z_k^i$ can also be estimated as part of the
Gibbs sampling procedure, see \cite{bore2017tracking}.

Since our learning scheme is EM, each iteration
of the algorithm should improve the expected
likelihood until arriving at a local maximum.
Given the particles at the last time step $K$,
the filter expected likelihood for the entire sequence is given by
\begin{equation*}
\mathbb{E} \left[ \mathcal{L}(\mathbf{Y}_{1:K} | c_{1:K}) \right] = \sum_i w_K^i \prod_{k=1}^K  \mathcal{L}(\mathbf{Y}_k | c_k).
\end{equation*}

\subsection{Target Clustering}
\label{sec:clustering}

With the described sampling process, we are able to propagate
the particles as new targets appear. However, we have yet to
describe the mechanism for combining the individual particles $i$
into a joint estimate of the number of targets $N_k$ and of their
states. First off, the number of targets can be easily estimated
by constructing a histogram from each sample's number of targets $N_k^i = \sum_j \mathds{1}_\text{alive}(e_{j, k}^i) $
and its weight.
Similar to e.g. \cite{kreucher2005multitarget}, the joint number can be estimated as the maximum of the weighted histogram over particle set sizes,
$$\tilde N_k = \arg\max_N \sum_i \mathds{1}_N(N_k^i) w_k^i.$$

In order to get a canonical assignment of particle target
identities to a joint collection of identities, we assign each
new point measurement $\mathbf{y}_{k, m}$ a target identity $j$. The idea is that a
measurement is assigned to one of the already alive target identities $j$,
if sufficiently many associations to that identity were sampled.
If not, we assign a new target identity $j$
to the measurement, with $e_{j, k-1}^i = \text{unborn}$ for all particles $i$.
Any target births associated with the
measurement will get the identity that was assigned to the
measurement. Note that this means that previously existing targets
can still be associated with the measurement without changing
target identity.
To find the measurement target identity $j$, we find the most common association
$$\kappa_m = \max_j \sum_i \mathds{1}_m(c_{j, k}^i) \mathds{1}_\text{alive}(e_{j, k-1}^i) w_k^i$$
for every measurement $m$.
If $\kappa_m  > \kappa$ with $\kappa$ being some threshold (0.5 in our experiments),
we assign measurement $m$ the target identity $j$ corresponding to $\kappa_m$.
Otherwise, we assign it a new target identity,
thus expanding the set of possible target births.
Every sampled birth associated with such a
measurement is estimated to be a new target.

Finally, we take the $\tilde N_k$ target ids with the largest weights
$\tilde{w}_{j, k} = \sum_i \mathds{1}_\text{alive}(e_{j, k}^i) w_k^i$
to be our new set of targets.
The states $(\mathbf{\hat{x}}_{j, k}, l_{j, k})$ can then be
estimated as normal, as the mean (for $\mathbf{\hat{x}}_{j, k}$) 
and mode (for $l_{j, k}$) of the weighted particle samples.

\subsection{Filtering Algorithm}
\label{sec:filtering_algorithm}

The filtering algorithm can be summarized in the following steps:
\begin{enumerate}
\item Sample assignments $c_k^i \sim q(c_k)$ using Gibbs sampling
\item Compute the weights $w_k^i$ in the same MCMC chain, see \cite{bore2017tracking}
\item Update the Kalman filter estimates $\boldsymbol{\mu}_{j, k}^i, \boldsymbol{\Sigma}_{j, k}^i$
\item Cluster the newborn targets $e_{j, k-1}^i = \text{unborn}, e_{j, k}^i = \text{alive}$
\item Estimate number of targets $\tilde N_k$ and posteriors
\end{enumerate}

\subsection{Smoothed Positions}
\label{sec:smoothing}

For estimation of the parameters of the local process,
we need maximum likelihood estimates of the positions and features using all available data.
One approach would be to extend the proposed filter to also
do smoothing, e.g. by sampling
particle trajectories as in \cite{sarkka2012backward}.
We have implemented such an approach but found that the complexity
grows significantly, while only marginally improving
our results. This is to be expected, since correctly inferring one
of the jumps with such a smoother requires that it has been sampled
by at least one filter particle.
To speed up the learning, we will instead assume that weighted 
associations sampled by the filter mirror the distribution conditioned on all data.
The idea is then to smooth the trajectories in between the
jumps, when the objects are in the same location, using a Kalman smoother.
Smoothing the trajectories is important in this context,
since a filter tends to underestimate the measurement noise.

In the following, we define the smoothing sequences for each
particle $i$ and target $j$. Note that these sequences are used in
the Kalman smoothing as well as for producing noise estimates.
The sequence of time steps where target $j$ jumps is given by
$$J_{i, j} = \left\langle k; u_{j, k}^i = \text{jump} \vee k = 0 \right\rangle.$$
For each of these jumps, we can also define the sequence of all time steps
where the target is at the estimated new position, and is associated
with an observation. All such time steps up to the next jump are given by:
$$S_{i, j}^n = \left\langle k; c_{j, k}^i \neq \epsilon, J_{i, j}(n) \leq k < J_{i, j}(n+1) \right\rangle.$$
These sequences are used somewhat interchangeably as sequences
and sets.
For each sampled continuous sequence 
$S_{i, j}^n$, we then use the already computed filtered position estimates
$\boldsymbol{\mu}_{j, k}^{s, i}, \boldsymbol{\Sigma}_{j, k}^{s, i}$
together with a backward pass of a \textit{Rauch-Tung-Striebel} (RTS)
smoother to get the smoothed estimates
$\tilde{\boldsymbol{\mu}}_{j, k}^{s, i}$,
$\tilde{\boldsymbol{\Sigma}}_{j, k}^{s, i}$.
Since we assume no process noise for the features,
the smoothed mean $\tilde{\boldsymbol{\mu}}_{j, k}^{f, i}$ 
is given by the mean of all associated
measurements over the full sequence $1, \dots, K$.
These estimates are used to update the Kalman process
parameters, as described in the following section.

\subsection{Parameter Estimation}
\label{sec:estimation}

Recall that we would like to estimate the dynamics of the
targets, both jointly for all objects, and for the
individual targets.
We use the framework of the EM algorithm to do this in an
iterative fashion. In each step, we want to calculate the
maximum likelihood estimates of the parameters 
$p_\text{jump}$, $\sigma_q$, and $\mathbf{R}^f$.

The number of jumps during the sequence can be modeled
as a binomial distribution, parameterized by $p_\text{jump}$.
Then, the maximum likelihood estimate of the parameter is
given by the number of positive events divided by the total
number of events, i.e. the number of jumps divided by the 
number of time steps that the object was alive,
$$p_\text{jump}^j = \frac{\sum_i \sum_{k=1}^K \mathds{1}_\text{alive}(e_{j, k}^i) \mathds{1}_\text{jump}(u_{j, k}^i) w_K^i }{\sum_i \sum_{k=1}^K \mathds{1}_\text{alive}(e_{j, k}^i) w_K^i}.$$

However, we have found that our approximate inference requires us to 
also take into account how many time steps $\delta$
that a location is typically unobserved.
Since the filter only samples a limited number of associations at
each step, sampling low-probability jumps from locations that were not
observed is highly unlikely. We therefore need to modify
$p_\text{jump}$ to include the unsampled jumps from
unobserved locations, by incorporating
the jump probabilities during a typical absence $\delta$.
With the mean absence away from an object given by
$$\tilde \delta = \frac{1}{\sum M_k} \sum_k M_k \min_{k'<k: l^y_{k'}=l^y_k} k - k',$$
our modified $p_\text{jump}$ estimate is (since $1-(1-p)^{\tilde \delta} \approx \tilde \delta p$ for small $p$)
$$\tilde p_\text{jump}^j = 1-(1-p_\text{jump}^j)^{\tilde \delta} \approx \tilde \delta \times p_\text{jump}^j.$$

To calculate the process covariance of the positions, we use a weighted
version of the estimate of the linear Gaussian case.
The EM algorithm for the linear Gaussian case was first described
by Schumway and Stoffer in \cite{shumway1982approach}.
In essence, it amounts to the covariance estimate of the
difference in estimated mean positions between subsequent time steps
plus some covariance to account for the estimated filter uncertainty. With
\begin{align*}
\mathbf{q}_{i, j}^*(k, k') &= (\tilde{\boldsymbol{\mu}}_{j, k}^{s, i} - \tilde{\boldsymbol{\mu}}_{j, k'}^{s, i})(\tilde{\boldsymbol{\mu}}_{j, k}^{s, i} - \tilde{\boldsymbol{\mu}}_{j, k'}^{s, i})^T \\
&+ \tilde{\boldsymbol{\Sigma}}_{j, k}^{s, i} + \tilde{\boldsymbol{\Sigma}}_{j, k'}^{s, i} - 2 cov(\mathbf{\hat{x}}_{j, k}^s, \mathbf{\hat{x}}_{j, k'}^s | \mathbf{Y}_{1:K}),
\end{align*}
we get the weighted process covariance estimate as
\begin{equation*}
\tilde{\mathbf{Q}}_j^s = \frac{\sum\limits_i \sum\limits_{n \in J_{i, j}} \:\: \sum\limits_{k = 1}^{|S_{i, j}^n|-1} \mathds{1}_\text{alive}(e_{j, k}^i) \mathbf{q}_{i, j}^*(S_{i, j}^n(k+1), S_{i, j}^n(k)) w_K^i}{\sum_i \sum_{n \in J_{i, j}} \sum_{k = 1}^{|S_{i, j}^n|-1} \mathds{1}_\text{alive}(e_{j, k}^i) w_K^i}.
\end{equation*}

The feature measurement noise is computed analogously, but 
using the differences between estimated feature positions 
and estimates instead. Since we have no process noise,
this amounts to the standard maximum likelihood estimate
of the covariance.
With
\begin{equation*}
\mathbf{r}_{i, j}^*(k) = (\mathbf{\hat{y}}_{k, m}^f - \tilde{\boldsymbol{\mu}}_{j, k}^{f, i})(\mathbf{\hat{y}}_{k, m}^f - \tilde{\boldsymbol{\mu}}_{j, k}^{f, i})^T,
\end{equation*}
it is given by the weighted covariance
\begin{equation*}
\tilde{\mathbf{R}}_j^f = \frac{\sum_i \sum_{n \in J_{i, j}} \sum_{k = 1}^{|S_{i, j}^n|} \mathds{1}_\text{alive}(e_{j, k}^i) \mathbf{r}_{i, j}^*(S_{i, j}^n(k)) w_K^i}{\sum_i \sum_{n \in J_{i, j}} \sum_{k = 1}^{|S_{i, j}^n|} \mathds{1}_\text{alive}(e_{j, k}^i) w_K^i}.
\end{equation*}

We may also estimate the parameters of all instances jointly.
For this purpose, we may view all inferred targets as belonging to the same instance.
The joint estimates are produced by summing the numerators and denominators
of the instance estimates separately, followed by a division of the
numerator and denominator sums.

\subsection{EM Algorithm}
\label{sec:em_algorithm}

Now we are ready to describe the EM iterations. The algorithm
is initialized with the parameters in Table \ref{table:parameters}.
The E-step is then performed by running the filter on one full
sequence of the given data. The filter estimates all the properties,
such as the number of jumps, needed for the subsequent M-step.
In the M-step, we estimate one or all of the three parameters
$p_\text{jump}$, $\sigma_q$, and $\mathbf{R}^f$,
as described in the results.
For the next iteration of the algorithm, we update the
estimated values in the filtering.
In all experiments, we run 10 iterations of the algorithm, which
proved enough for the parameter estimates to converge.
In summary, the learning includes the steps:
\begin{enumerate}
\item Sample $c_k^i, \boldsymbol{\mu}_{j, k}^i, \boldsymbol{\Sigma}_{j, k}^i$ with the filter from Section \ref{sec:inference}
\item Smooth the continuous object trajectories to get $\tilde{\boldsymbol{\mu}}_{j, k}^i, \tilde{\boldsymbol{\Sigma}}_{j, k}^i$ using the scheme of Section \ref{sec:smoothing}
\item Given these parameters and $c_k^i$, estimate $p_\text{jump}, \sigma_q, \mathbf{R}^f$ using the method of Section \ref{sec:estimation}
\item Repeat with the new parameter estimates
\end{enumerate}

%% file: prior_table.tex
\definecolor{lightgray}{gray}{0.9}

\begin{table*}[htpb]
\begin{center}
\rowcolors{5}{}{lightgray}
\begin{tabular}{c|rrrrrr}
  \scalebox{1}[-1]{$\hookrightarrow$} & $e_{j, k} = \text{alive},$ & $e_{j, k} = \text{alive},$ & $e_{j, k} = \text{alive},$ & $e_{j, k} = \text{alive},$ &  $e_{j, k} = \text{alive},$ & \multirow{4}{*}{$e_{j, k} = \text{dead}$} \\
  $p(e_{j, k}, u_{j, k}, l_{j, k}, c_{j, k} \: | $ & $u_{j, k} = \text{no jump},$ & $u_{j, k} = \text{no jump},$ & $u_{j, k} = \text{jump},$ & $u_{j, k} = \text{jump},$ & $u_{j, k} = \text{jump},$ & \\
  $e_{j, k-1}, l_{j, k-1})$ & $l_{j, k} = l_{j, k-1},$ & $l_{j, k} = l_{j, k-1},$ & $l_{j, k} = l_k^y,$ & $l_{j, k} = l_k^y,$ & $l_{j, k} = l_\text{unknown},$ & \\
  $\downarrow$ & $c_{j,k} = m$ & $c_{j,k} = \epsilon$ & $c_{j,k} = m$ & $c_{j,k} = \epsilon$ & $c_{j,k} = \epsilon$ & \\
  \hline
  \rowcolor{lightgray} & $\frac{1}{M_k} p_{\text{life}} \cdot $ & $p_{\text{life}} (1-p_{\text{jump}}) \cdot$ & & & & \\
  \rowcolor{lightgray}\multirow{-2}{*}{$e_{j, k-1} = \text{alive}, l_{j, k-1} = l_k^y$} & $(1-p_{\text{jump}})p_{\text{meas}}$ & $(1-p_{\text{meas}})$ & \multirow{-2}{*}{$\frac{1}{M_k N_l} p_{\text{life}}p_{\text{jump}}p_{\text{meas}}$} &  \multirow{-2}{*}{$\frac{1}{N_l} p_{\text{life}}p_{\text{jump}}(1-p_{\text{meas}})$} & \multirow{-2}{*}{$\frac{N_l-1}{N_l}p_{\text{life}}p_{\text{jump}}$} & \multirow{-2}{*}{$p_{\text{death}}$} \\
  $e_{j, k-1} = \text{alive}, l_{j, k-1} \neq l_k^y$ & 0 & $p_{\text{life}}(1-p_{\text{jump}})$ & $\frac{1}{M_k N_l}p_{\text{life}}p_{\text{jump}}p_{\text{meas}}$ & $\frac{1}{N_l}p_{\text{life}}p_{\text{jump}}(1-p_{\text{meas}})$ & $\frac{N_l-1}{N_l}p_{\text{life}}p_{\text{jump}}$ & $p_{\text{death}}$ \\
  $e_{j, k-1} = \text{alive}, l_{j, k-1} = l_\text{unknown}$ & 0 & 0 & $\frac{1}{M_k N_l}p_{\text{life}}p_{\text{meas}}$ & $\frac{1}{N_l}p_{\text{life}}(1-p_{\text{meas}})$ & $\frac{N_l-1}{N_l}p_{\text{life}}$ & $p_{\text{death}}$ \\
  $e_{j, k-1} = \text{unborn}$ & $\frac{1}{M_k} p_{\text{birth}}$ & 0 & 0 & 0 & 0 & $1-p_{\text{birth}}$ \\
\end{tabular}
\end{center}
\caption{Prior probabilities given existence and previous location.
         The top 3 rows, and 5 leftmost columns correspond to the
         transition prior $p(u_{j, k}, l_{j, k}, c_{j, k} | l_{j, k-1})$ 
         of a target that is always alive (see Section \ref{sec:prior}).
         $l_\text{unknown}$ (5:th column) signifies any location that the target jumped to
         that is not the currently observed location $l^y_k$,
         see \cite{bore2017tracking} for details. In our proposal
         sampling (see Section \ref{sec:inference}), we may sample $M_k$
         previously unborn associations, one for each measurement. For the most part,
         those will sample $e_{j, k} = \text{dead}$, yielding no new targets.}
\label{table:prior}
\end{table*}

%% file: img/model.tex
\ifx\du\undefined
  \newlength{\du}
\fi
\setlength{\du}{8.5\unitlength}
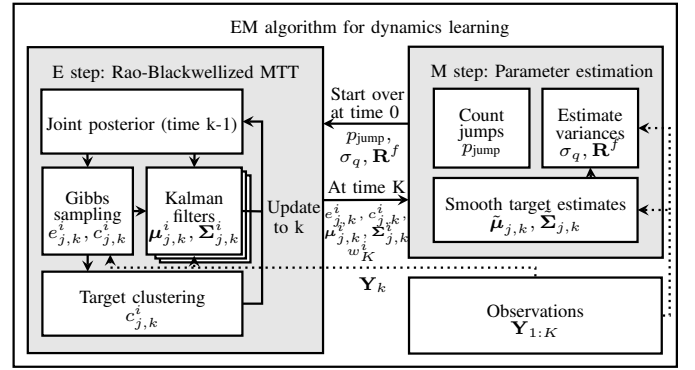
\begin{figure}
\begin{tikzpicture}
\scriptsize
\pgftransformxscale{1.000000}
\pgftransformyscale{-1.000000}
\definecolor{dialinecolor}{rgb}{0.000000, 0.000000, 0.000000}
\pgfsetstrokecolor{dialinecolor}
\definecolor{dialinecolor}{rgb}{1.000000, 1.000000, 1.000000}
\pgfsetfillcolor{dialinecolor}
\definecolor{dialinecolor}{rgb}{1.000000, 1.000000, 1.000000}
\pgfsetfillcolor{dialinecolor}
\fill (15.114662\du,7.842110\du)--(15.114662\du,24.008250\du)--(44.629954\du,24.008250\du)--(44.629954\du,7.842110\du)--cycle;
\pgfsetlinewidth{0.100000\du}
\pgfsetdash{}{0pt}
\pgfsetdash{}{0pt}
\pgfsetmiterjoin
\definecolor{dialinecolor}{rgb}{0.000000, 0.000000, 0.000000}
\pgfsetstrokecolor{dialinecolor}
\draw (15.114662\du,7.842110\du)--(15.114662\du,24.008250\du)--(44.629954\du,24.008250\du)--(44.629954\du,7.842110\du)--cycle;
\definecolor{dialinecolor}{rgb}{0.000000, 0.000000, 0.000000}
\pgfsetstrokecolor{dialinecolor}
\node at (29.872308\du,16.120180\du){};
\definecolor{dialinecolor}{rgb}{0.000000, 0.000000, 0.000000}
\pgfsetstrokecolor{dialinecolor}
\node[anchor=west] at (24.443236\du,8.967006\du){EM algorithm for dynamics learning};
\definecolor{dialinecolor}{rgb}{0.9, 0.9, 0.9}
\pgfsetfillcolor{dialinecolor}
\fill (15.745330\du,9.775010\du)--(15.745330\du,23.398603\du)--(28.884060\du,23.398603\du)--(28.884060\du,9.775010\du)--cycle;
\pgfsetlinewidth{0.100000\du}
\pgfsetdash{}{0pt}
\pgfsetdash{}{0pt}
\pgfsetmiterjoin
\definecolor{dialinecolor}{rgb}{0.000000, 0.000000, 0.000000}
\pgfsetstrokecolor{dialinecolor}
\draw (15.745330\du,9.775010\du)--(15.745330\du,23.398603\du)--(28.884060\du,23.398603\du)--(28.884060\du,9.775010\du)--cycle;
\definecolor{dialinecolor}{rgb}{0.000000, 0.000000, 0.000000}
\pgfsetstrokecolor{dialinecolor}
\node at (22.314695\du,16.781807\du){};
\definecolor{dialinecolor}{rgb}{0.000000, 0.000000, 0.000000}
\pgfsetstrokecolor{dialinecolor}
\node[anchor=west] at (16.527348\du,10.972166\du){E step: Rao-Blackwellized MTT};
\definecolor{dialinecolor}{rgb}{0.000000, 0.000000, 0.000000}
\pgfsetstrokecolor{dialinecolor}
\node[anchor=west] at (21.300100\du,10.825010\du){};
\definecolor{dialinecolor}{rgb}{0.9, 0.9, 0.9}
\pgfsetfillcolor{dialinecolor}
\fill (32.700526\du,9.755138\du)--(32.700526\du,19.152101\du)--(44.000526\du,19.152101\du)--(44.000526\du,9.755138\du)--cycle;
\pgfsetlinewidth{0.100000\du}
\pgfsetdash{}{0pt}
\pgfsetdash{}{0pt}
\pgfsetmiterjoin
\definecolor{dialinecolor}{rgb}{0.000000, 0.000000, 0.000000}
\pgfsetstrokecolor{dialinecolor}
\draw (32.700526\du,9.755138\du)--(32.700526\du,19.152101\du)--(44.000526\du,19.152101\du)--(44.000526\du,9.755138\du)--cycle;
\definecolor{dialinecolor}{rgb}{0.000000, 0.000000, 0.000000}
\pgfsetstrokecolor{dialinecolor}
\node at (38.350526\du,14.648620\du){};
\definecolor{dialinecolor}{rgb}{0.000000, 0.000000, 0.000000}
\pgfsetstrokecolor{dialinecolor}
\node[anchor=west] at (33.350526\du,10.920010\du){M step: Parameter estimation};
\definecolor{dialinecolor}{rgb}{0.000000, 0.000000, 0.000000}
\pgfsetstrokecolor{dialinecolor}
\node[anchor=west] at (33.895526\du,10.725010\du){};
\definecolor{dialinecolor}{rgb}{0.000000, 0.000000, 0.000000}
\pgfsetstrokecolor{dialinecolor}
\node[anchor=west] at (27.850100\du,6.875010\du){};
\definecolor{dialinecolor}{rgb}{1.000000, 1.000000, 1.000000}
\pgfsetfillcolor{dialinecolor}
\fill (16.351236\du,11.662532\du)--(16.351236\du,14.512528\du)--(25.298736\du,14.512528\du)--(25.298736\du,11.662532\du)--cycle;
\pgfsetlinewidth{0.100000\du}
\pgfsetdash{}{0pt}
\pgfsetdash{}{0pt}
\pgfsetmiterjoin
\definecolor{dialinecolor}{rgb}{0.000000, 0.000000, 0.000000}
\pgfsetstrokecolor{dialinecolor}
\draw (16.351236\du,11.662532\du)--(16.351236\du,14.512528\du)--(25.298736\du,14.512528\du)--(25.298736\du,11.662532\du)--cycle;
\definecolor{dialinecolor}{rgb}{0.000000, 0.000000, 0.000000}
\pgfsetstrokecolor{dialinecolor}
\node at (20.824986\du,13.282530\du){Joint posterior (time k-1)};
\definecolor{dialinecolor}{rgb}{1.000000, 1.000000, 1.000000}
\pgfsetfillcolor{dialinecolor}
\fill (16.399994\du,15.137526\du)--(16.399994\du,19.012519\du)--(20.422494\du,19.012519\du)--(20.422494\du,15.137526\du)--cycle;
\pgfsetlinewidth{0.100000\du}
\pgfsetdash{}{0pt}
\pgfsetdash{}{0pt}
\pgfsetmiterjoin
\definecolor{dialinecolor}{rgb}{0.000000, 0.000000, 0.000000}
\pgfsetstrokecolor{dialinecolor}
\draw (16.399994\du,15.137526\du)--(16.399994\du,19.012519\du)--(20.422494\du,19.012519\du)--(20.422494\du,15.137526\du)--cycle;
\definecolor{dialinecolor}{rgb}{0.000000, 0.000000, 0.000000}
\pgfsetstrokecolor{dialinecolor}
\node at (18.411244\du,16.470022\du){Gibbs};
\definecolor{dialinecolor}{rgb}{0.000000, 0.000000, 0.000000}
\pgfsetstrokecolor{dialinecolor}
\node at (18.411244\du,17.270022\du){sampling};
\definecolor{dialinecolor}{rgb}{0.000000, 0.000000, 0.000000}
\pgfsetstrokecolor{dialinecolor}
\node at (18.411244\du,18.070022\du){$e_{j, k}^i, c_{j, k}^i$};
\definecolor{dialinecolor}{rgb}{1.000000, 1.000000, 1.000000}
\pgfsetfillcolor{dialinecolor}
\fill (21.627484\du,15.485026\du)--(21.627484\du,19.360019\du)--(25.654984\du,19.360019\du)--(25.654984\du,15.485026\du)--cycle;
\pgfsetlinewidth{0.100000\du}
\pgfsetdash{}{0pt}
\pgfsetdash{}{0pt}
\pgfsetmiterjoin
\definecolor{dialinecolor}{rgb}{0.000000, 0.000000, 0.000000}
\pgfsetstrokecolor{dialinecolor}
\draw (21.627484\du,15.485026\du)--(21.627484\du,19.360019\du)--(25.654984\du,19.360019\du)--(25.654984\du,15.485026\du)--cycle;
\definecolor{dialinecolor}{rgb}{0.000000, 0.000000, 0.000000}
\pgfsetstrokecolor{dialinecolor}
\node at (23.641234\du,17.617523\du){};
\definecolor{dialinecolor}{rgb}{1.000000, 1.000000, 1.000000}
\pgfsetfillcolor{dialinecolor}
\fill (21.424985\du,15.310026\du)--(21.424985\du,19.185019\du)--(25.452485\du,19.185019\du)--(25.452485\du,15.310026\du)--cycle;
\pgfsetlinewidth{0.100000\du}
\pgfsetdash{}{0pt}
\pgfsetdash{}{0pt}
\pgfsetmiterjoin
\definecolor{dialinecolor}{rgb}{0.000000, 0.000000, 0.000000}
\pgfsetstrokecolor{dialinecolor}
\draw (21.424985\du,15.310026\du)--(21.424985\du,19.185019\du)--(25.452485\du,19.185019\du)--(25.452485\du,15.310026\du)--cycle;
\definecolor{dialinecolor}{rgb}{0.000000, 0.000000, 0.000000}
\pgfsetstrokecolor{dialinecolor}
\node at (23.438735\du,17.442523\du){};
\definecolor{dialinecolor}{rgb}{1.000000, 1.000000, 1.000000}
\pgfsetfillcolor{dialinecolor}
\fill (21.034033\du,15.135027\du)--(21.034033\du,19.010020\du)--(25.268227\du,19.010020\du)--(25.268227\du,15.135027\du)--cycle;
\pgfsetlinewidth{0.100000\du}
\pgfsetdash{}{0pt}
\pgfsetdash{}{0pt}
\pgfsetmiterjoin
\definecolor{dialinecolor}{rgb}{0.000000, 0.000000, 0.000000}
\pgfsetstrokecolor{dialinecolor}
\draw (21.034033\du,15.135027\du)--(21.034033\du,19.010020\du)--(25.268227\du,19.010020\du)--(25.268227\du,15.135027\du)--cycle;
\definecolor{dialinecolor}{rgb}{0.000000, 0.000000, 0.000000}
\pgfsetstrokecolor{dialinecolor}
\node at (23.151130\du,16.467523\du){Kalman};
\definecolor{dialinecolor}{rgb}{0.000000, 0.000000, 0.000000}
\pgfsetstrokecolor{dialinecolor}
\node at (23.151130\du,17.267523\du){filters};
\definecolor{dialinecolor}{rgb}{0.000000, 0.000000, 0.000000}
\pgfsetstrokecolor{dialinecolor}
\node at (23.151130\du,18.067523\du){$\boldsymbol{\mu}_{j, k}^i, \boldsymbol{\Sigma}_{j, k}^i$};
\definecolor{dialinecolor}{rgb}{0.000000, 0.000000, 0.000000}
\pgfsetstrokecolor{dialinecolor}
\node[anchor=west] at (23.151130\du,17.072523\du){};
\definecolor{dialinecolor}{rgb}{0.000000, 0.000000, 0.000000}
\pgfsetstrokecolor{dialinecolor}
\node[anchor=west] at (23.438735\du,17.247523\du){};
\definecolor{dialinecolor}{rgb}{0.000000, 0.000000, 0.000000}
\pgfsetstrokecolor{dialinecolor}
\node[anchor=west] at (23.151130\du,17.072523\du){};
\definecolor{dialinecolor}{rgb}{1.000000, 1.000000, 1.000000}
\pgfsetfillcolor{dialinecolor}
\fill (16.395902\du,19.772881\du)--(16.395902\du,22.622876\du)--(25.295884\du,22.622876\du)--(25.295884\du,19.772881\du)--cycle;
\pgfsetlinewidth{0.100000\du}
\pgfsetdash{}{0pt}
\pgfsetdash{}{0pt}
\pgfsetmiterjoin
\definecolor{dialinecolor}{rgb}{0.000000, 0.000000, 0.000000}
\pgfsetstrokecolor{dialinecolor}
\draw (16.395902\du,19.772881\du)--(16.395902\du,22.622876\du)--(25.295884\du,22.622876\du)--(25.295884\du,19.772881\du)--cycle;
\definecolor{dialinecolor}{rgb}{0.000000, 0.000000, 0.000000}
\pgfsetstrokecolor{dialinecolor}
\node at (20.845893\du,20.992879\du){Target clustering};
\definecolor{dialinecolor}{rgb}{0.000000, 0.000000, 0.000000}
\pgfsetstrokecolor{dialinecolor}
\node at (20.845893\du,21.792879\du){$c_{j, k}^i$};
\definecolor{dialinecolor}{rgb}{0.000000, 0.000000, 0.000000}
\pgfsetstrokecolor{dialinecolor}
\node[anchor=west] at (20.845893\du,21.197879\du){};
\definecolor{dialinecolor}{rgb}{0.000000, 0.000000, 0.000000}
\pgfsetstrokecolor{dialinecolor}
\node[anchor=west] at (20.845893\du,21.197879\du){};
\definecolor{dialinecolor}{rgb}{1.000000, 1.000000, 1.000000}
\pgfsetfillcolor{dialinecolor}
\fill (33.810648\du,15.600064\du)--(33.810648\du,18.450059\du)--(42.880648\du,18.450059\du)--(42.880648\du,15.600064\du)--cycle;
\pgfsetlinewidth{0.100000\du}
\pgfsetdash{}{0pt}
\pgfsetdash{}{0pt}
\pgfsetmiterjoin
\definecolor{dialinecolor}{rgb}{0.000000, 0.000000, 0.000000}
\pgfsetstrokecolor{dialinecolor}
\draw (33.810648\du,15.600064\du)--(33.810648\du,18.450059\du)--(42.880648\du,18.450059\du)--(42.880648\du,15.600064\du)--cycle;
\definecolor{dialinecolor}{rgb}{0.000000, 0.000000, 0.000000}
\pgfsetstrokecolor{dialinecolor}
\node at (38.345648\du,16.820062\du){Smooth target estimates};
\definecolor{dialinecolor}{rgb}{0.000000, 0.000000, 0.000000}
\pgfsetstrokecolor{dialinecolor}
\node at (38.345648\du,17.620062\du){$\tilde{\boldsymbol{\mu}}_{j, k}, \tilde{\boldsymbol{\Sigma}}_{j, k}$};
\definecolor{dialinecolor}{rgb}{0.000000, 0.000000, 0.000000}
\pgfsetstrokecolor{dialinecolor}
\node[anchor=west] at (38.345648\du,17.025062\du){};
\definecolor{dialinecolor}{rgb}{1.000000, 1.000000, 1.000000}
\pgfsetfillcolor{dialinecolor}
\fill (33.814541\du,11.626122\du)--(33.814541\du,15.137137\du)--(38.069450\du,15.137137\du)--(38.069450\du,11.626122\du)--cycle;
\pgfsetlinewidth{0.100000\du}
\pgfsetdash{}{0pt}
\pgfsetdash{}{0pt}
\pgfsetmiterjoin
\definecolor{dialinecolor}{rgb}{0.000000, 0.000000, 0.000000}
\pgfsetstrokecolor{dialinecolor}
\draw (33.814541\du,11.626122\du)--(33.814541\du,15.137137\du)--(38.069450\du,15.137137\du)--(38.069450\du,11.626122\du)--cycle;
\definecolor{dialinecolor}{rgb}{0.000000, 0.000000, 0.000000}
\pgfsetstrokecolor{dialinecolor}
\node at (35.941995\du,12.776630\du){Count};
\definecolor{dialinecolor}{rgb}{0.000000, 0.000000, 0.000000}
\pgfsetstrokecolor{dialinecolor}
\node at (35.941995\du,13.576630\du){jumps};
\definecolor{dialinecolor}{rgb}{0.000000, 0.000000, 0.000000}
\pgfsetstrokecolor{dialinecolor}
\node at (35.941995\du,14.376630\du){$p_\text{jump}$};
\definecolor{dialinecolor}{rgb}{0.000000, 0.000000, 0.000000}
\pgfsetstrokecolor{dialinecolor}
\node[anchor=west] at (35.941995\du,13.381630\du){};
\definecolor{dialinecolor}{rgb}{1.000000, 1.000000, 1.000000}
\pgfsetfillcolor{dialinecolor}
\fill (38.648856\du,11.627796\du)--(38.648856\du,15.127796\du)--(42.903765\du,15.127796\du)--(42.903765\du,11.627796\du)--cycle;
\pgfsetlinewidth{0.100000\du}
\pgfsetdash{}{0pt}
\pgfsetdash{}{0pt}
\pgfsetmiterjoin
\definecolor{dialinecolor}{rgb}{0.000000, 0.000000, 0.000000}
\pgfsetstrokecolor{dialinecolor}
\draw (38.648856\du,11.627796\du)--(38.648856\du,15.127796\du)--(42.903765\du,15.127796\du)--(42.903765\du,11.627796\du)--cycle;
\definecolor{dialinecolor}{rgb}{0.000000, 0.000000, 0.000000}
\pgfsetstrokecolor{dialinecolor}
\node at (40.776310\du,12.772796\du){Estimate};
\definecolor{dialinecolor}{rgb}{0.000000, 0.000000, 0.000000}
\pgfsetstrokecolor{dialinecolor}
\node at (40.776310\du,13.572796\du){variances};
\definecolor{dialinecolor}{rgb}{0.000000, 0.000000, 0.000000}
\pgfsetstrokecolor{dialinecolor}
\node at (40.776310\du,14.372796\du){$\sigma_q, \mathbf{R}^f$};
\pgfsetlinewidth{0.100000\du}
\pgfsetdash{}{0pt}
\pgfsetdash{}{0pt}
\pgfsetbuttcap
{
\definecolor{dialinecolor}{rgb}{0.000000, 0.000000, 0.000000}
\pgfsetfillcolor{dialinecolor}
\pgfsetarrowsend{stealth}
\definecolor{dialinecolor}{rgb}{0.000000, 0.000000, 0.000000}
\pgfsetstrokecolor{dialinecolor}
\draw (18.398910\du,14.512528\du)--(18.411244\du,15.137526\du);
}
\pgfsetlinewidth{0.100000\du}
\pgfsetdash{}{0pt}
\pgfsetdash{}{0pt}
\pgfsetbuttcap
{
\definecolor{dialinecolor}{rgb}{0.000000, 0.000000, 0.000000}
\pgfsetfillcolor{dialinecolor}
\pgfsetarrowsend{stealth}
\definecolor{dialinecolor}{rgb}{0.000000, 0.000000, 0.000000}
\pgfsetstrokecolor{dialinecolor}
\draw (23.166972\du,14.512528\du)--(23.151130\du,15.135027\du);
}
\pgfsetlinewidth{0.100000\du}
\pgfsetdash{}{0pt}
\pgfsetdash{}{0pt}
\pgfsetbuttcap
{
\definecolor{dialinecolor}{rgb}{0.000000, 0.000000, 0.000000}
\pgfsetfillcolor{dialinecolor}
\pgfsetarrowsend{stealth}
\definecolor{dialinecolor}{rgb}{0.000000, 0.000000, 0.000000}
\pgfsetstrokecolor{dialinecolor}
\draw (20.422494\du,17.075022\du)--(21.034033\du,17.072523\du);
}
\pgfsetlinewidth{0.100000\du}
\pgfsetdash{}{0pt}
\pgfsetdash{}{0pt}
\pgfsetbuttcap
{
\definecolor{dialinecolor}{rgb}{0.000000, 0.000000, 0.000000}
\pgfsetfillcolor{dialinecolor}
\pgfsetarrowsend{stealth}
\definecolor{dialinecolor}{rgb}{0.000000, 0.000000, 0.000000}
\pgfsetstrokecolor{dialinecolor}
\draw (18.411244\du,19.012519\du)--(18.410674\du,19.772881\du);
}
\pgfsetlinewidth{0.100000\du}
\pgfsetdash{}{0pt}
\pgfsetdash{}{0pt}
\pgfsetmiterjoin
\pgfsetbuttcap
{
\definecolor{dialinecolor}{rgb}{0.000000, 0.000000, 0.000000}
\pgfsetfillcolor{dialinecolor}
\pgfsetarrowsend{stealth}
{\pgfsetcornersarced{\pgfpoint{0.000000\du}{0.000000\du}}\definecolor{dialinecolor}{rgb}{0.000000, 0.000000, 0.000000}
\pgfsetstrokecolor{dialinecolor}
\draw (25.295884\du,21.197879\du)--(26.172185\du,21.197879\du)--(26.172185\du,13.087530\du)--(25.298736\du,13.087530\du);
}}
\pgfsetlinewidth{0.100000\du}
\pgfsetdash{}{0pt}
\pgfsetdash{}{0pt}
\pgfsetbuttcap
{
\definecolor{dialinecolor}{rgb}{0.000000, 0.000000, 0.000000}
\pgfsetfillcolor{dialinecolor}
\definecolor{dialinecolor}{rgb}{0.000000, 0.000000, 0.000000}
\pgfsetstrokecolor{dialinecolor}
\draw (25.268227\du,17.072523\du)--(26.172185\du,17.089207\du);
}
\definecolor{dialinecolor}{rgb}{0.000000, 0.000000, 0.000000}
\pgfsetstrokecolor{dialinecolor}
\node[anchor=west] at (25.205160\du,10.530253\du){};
\definecolor{dialinecolor}{rgb}{0.000000, 0.000000, 0.000000}
\pgfsetstrokecolor{dialinecolor}
\node[anchor=west] at (22.314695\du,16.586807\du){};
\definecolor{dialinecolor}{rgb}{0.000000, 0.000000, 0.000000}
\pgfsetstrokecolor{dialinecolor}
\node[anchor=west] at (26.073974\du,16.955336\du){Update};
\definecolor{dialinecolor}{rgb}{0.000000, 0.000000, 0.000000}
\pgfsetstrokecolor{dialinecolor}
\node[anchor=west] at (26.373974\du,17.755336\du){to k};
\definecolor{dialinecolor}{rgb}{0.000000, 0.000000, 0.000000}
\pgfsetstrokecolor{dialinecolor}
\node[anchor=west] at (30.019264\du,16.164226\du){};
\definecolor{dialinecolor}{rgb}{0.000000, 0.000000, 0.000000}
\pgfsetstrokecolor{dialinecolor}
\node[anchor=west] at (23.151130\du,17.072523\du){};
\definecolor{dialinecolor}{rgb}{0.000000, 0.000000, 0.000000}
\pgfsetstrokecolor{dialinecolor}
\node[anchor=west] at (18.411244\du,17.075022\du){};
\pgfsetlinewidth{0.100000\du}
\pgfsetdash{}{0pt}
\pgfsetdash{}{0pt}
\pgfsetbuttcap
{
\definecolor{dialinecolor}{rgb}{0.000000, 0.000000, 0.000000}
\pgfsetfillcolor{dialinecolor}
\pgfsetarrowsend{stealth}
\definecolor{dialinecolor}{rgb}{0.000000, 0.000000, 0.000000}
\pgfsetstrokecolor{dialinecolor}
\draw (28.884060\du,16.586807\du)--(32.700526\du,16.571615\du);
}
\pgfsetlinewidth{0.100000\du}
\pgfsetdash{}{0pt}
\pgfsetdash{}{0pt}
\pgfsetbuttcap
{
\definecolor{dialinecolor}{rgb}{0.000000, 0.000000, 0.000000}
\pgfsetfillcolor{dialinecolor}
\pgfsetarrowsend{stealth}
\definecolor{dialinecolor}{rgb}{0.000000, 0.000000, 0.000000}
\pgfsetstrokecolor{dialinecolor}
\draw (32.689295\du,13.181771\du)--(28.884060\du,13.180908\du);
}
\definecolor{dialinecolor}{rgb}{0.000000, 0.000000, 0.000000}
\pgfsetstrokecolor{dialinecolor}
\node[anchor=west] at (28.775050\du,16.008756\du){At time K};
\definecolor{dialinecolor}{rgb}{0.000000, 0.000000, 0.000000}
\pgfsetstrokecolor{dialinecolor}
\node[anchor=west] at (20.824986\du,13.087530\du){};
\definecolor{dialinecolor}{rgb}{0.000000, 0.000000, 0.000000}
\pgfsetstrokecolor{dialinecolor}
\node[anchor=west] at (38.345648\du,17.025062\du){};
\definecolor{dialinecolor}{rgb}{0.000000, 0.000000, 0.000000}
\pgfsetstrokecolor{dialinecolor}
\node[anchor=west] at (28.901697\du,11.834931\du){Start over};
\definecolor{dialinecolor}{rgb}{0.000000, 0.000000, 0.000000}
\pgfsetstrokecolor{dialinecolor}
\node[anchor=west] at (28.901697\du,12.634931\du){at time 0};
\definecolor{dialinecolor}{rgb}{1.000000, 1.000000, 1.000000}
\pgfsetfillcolor{dialinecolor}
\fill (32.710317\du,20.035037\du)--(32.710317\du,23.419626\du)--(43.957241\du,23.419626\du)--(43.957241\du,20.035037\du)--cycle;
\pgfsetlinewidth{0.100000\du}
\pgfsetdash{}{0pt}
\pgfsetdash{}{0pt}
\pgfsetmiterjoin
\definecolor{dialinecolor}{rgb}{0.000000, 0.000000, 0.000000}
\pgfsetstrokecolor{dialinecolor}
\draw (32.710317\du,20.035037\du)--(32.710317\du,23.419626\du)--(43.957241\du,23.419626\du)--(43.957241\du,20.035037\du)--cycle;
\definecolor{dialinecolor}{rgb}{0.000000, 0.000000, 0.000000}
\pgfsetstrokecolor{dialinecolor}
\node at (38.333779\du,21.522332\du){Observations};
\definecolor{dialinecolor}{rgb}{0.000000, 0.000000, 0.000000}
\pgfsetstrokecolor{dialinecolor}
\node at (38.333779\du,22.322332\du){$\mathbf{Y}_{1:K}$};
\pgfsetlinewidth{0.100000\du}
\pgfsetdash{{\pgflinewidth}{0.200000\du}}{0cm}
\pgfsetdash{{\pgflinewidth}{0.200000\du}}{0cm}
\pgfsetmiterjoin
\pgfsetbuttcap
{
\definecolor{dialinecolor}{rgb}{0.000000, 0.000000, 0.000000}
\pgfsetfillcolor{dialinecolor}
\pgfsetarrowsend{stealth}
{\pgfsetcornersarced{\pgfpoint{0.000000\du}{0.000000\du}}\definecolor{dialinecolor}{rgb}{0.000000, 0.000000, 0.000000}
\pgfsetstrokecolor{dialinecolor}
\draw (38.333779\du,19.984794\du)--(38.333779\du,19.593569\du)--(23.151130\du,19.593569\du)--(23.151130\du,19.059940\du);
}}
\pgfsetlinewidth{0.100000\du}
\pgfsetdash{{\pgflinewidth}{0.200000\du}}{0cm}
\pgfsetdash{{\pgflinewidth}{0.200000\du}}{0cm}
\pgfsetmiterjoin
\pgfsetbuttcap
{
\definecolor{dialinecolor}{rgb}{0.000000, 0.000000, 0.000000}
\pgfsetfillcolor{dialinecolor}
\pgfsetarrowsend{stealth}
{\pgfsetcornersarced{\pgfpoint{0.000000\du}{0.000000\du}}\definecolor{dialinecolor}{rgb}{0.000000, 0.000000, 0.000000}
\pgfsetstrokecolor{dialinecolor}
\draw (23.124154\du,19.593569\du)--(19.416869\du,19.593569\du)--(19.416869\du,19.012519\du);
}}
\definecolor{dialinecolor}{rgb}{0.000000, 0.000000, 0.000000}
\pgfsetstrokecolor{dialinecolor}
\node[anchor=west] at (30.147700\du,20.354645\du){$\mathbf{Y}_k$};
\pgfsetlinewidth{0.100000\du}
\pgfsetdash{}{0pt}
\pgfsetdash{}{0pt}
\pgfsetbuttcap
{
\definecolor{dialinecolor}{rgb}{0.000000, 0.000000, 0.000000}
\pgfsetfillcolor{dialinecolor}
\pgfsetarrowsend{stealth}
\definecolor{dialinecolor}{rgb}{0.000000, 0.000000, 0.000000}
\pgfsetstrokecolor{dialinecolor}
\draw (40.802349\du,15.600064\du)--(40.776310\du,15.127796\du);
}
\pgfsetlinewidth{0.100000\du}
\pgfsetdash{{\pgflinewidth}{0.200000\du}}{0cm}
\pgfsetdash{{\pgflinewidth}{0.200000\du}}{0cm}
\pgfsetmiterjoin
\pgfsetbuttcap
{
\definecolor{dialinecolor}{rgb}{0.000000, 0.000000, 0.000000}
\pgfsetfillcolor{dialinecolor}
\pgfsetarrowsend{stealth}
{\pgfsetcornersarced{\pgfpoint{0.000000\du}{0.000000\du}}\definecolor{dialinecolor}{rgb}{0.000000, 0.000000, 0.000000}
\pgfsetstrokecolor{dialinecolor}
\draw (44.007243\du,21.727332\du)--(44.308840\du,21.727332\du)--(44.308840\du,17.025062\du)--(42.880648\du,17.025062\du);
}}
\pgfsetlinewidth{0.100000\du}
\pgfsetdash{{\pgflinewidth}{0.200000\du}}{0cm}
\pgfsetdash{{\pgflinewidth}{0.200000\du}}{0cm}
\pgfsetmiterjoin
\pgfsetbuttcap
{
\definecolor{dialinecolor}{rgb}{0.000000, 0.000000, 0.000000}
\pgfsetfillcolor{dialinecolor}
\pgfsetarrowsend{stealth}
{\pgfsetcornersarced{\pgfpoint{0.000000\du}{0.000000\du}}\definecolor{dialinecolor}{rgb}{0.000000, 0.000000, 0.000000}
\pgfsetstrokecolor{dialinecolor}
\draw (44.308840\du,17.025062\du)--(44.308840\du,13.377796\du)--(42.903765\du,13.377796\du);
}}
\definecolor{dialinecolor}{rgb}{0.000000, 0.000000, 0.000000}
\pgfsetstrokecolor{dialinecolor}
\node[anchor=west] at (31.422178\du,8.851180\du){};
\definecolor{dialinecolor}{rgb}{0.000000, 0.000000, 0.000000}
\pgfsetstrokecolor{dialinecolor}
\node[anchor=west] at (29.595341\du,13.837736\du){$p_\text{jump},$};
\definecolor{dialinecolor}{rgb}{0.000000, 0.000000, 0.000000}
\pgfsetstrokecolor{dialinecolor}
\node[anchor=west] at (29.295341\du,14.637736\du){$\sigma_q, \mathbf{R}^f$};
\definecolor{dialinecolor}{rgb}{0.000000, 0.000000, 0.000000}
\pgfsetstrokecolor{dialinecolor}
\node[anchor=west] at (30.770487\du,13.833462\du){};
\definecolor{dialinecolor}{rgb}{0.000000, 0.000000, 0.000000}
\pgfsetstrokecolor{dialinecolor}
\node[anchor=west] at (28.616363\du,17.264369\du){\tiny $e_{j, k}^i, c_{j, k}^i,$};
\definecolor{dialinecolor}{rgb}{0.000000, 0.000000, 0.000000}
\pgfsetstrokecolor{dialinecolor}
\node[anchor=west] at (28.616363\du,18.064369\du){\tiny $\boldsymbol{\mu}_{j, k}^i, \boldsymbol{\Sigma}_{j, k}^i$};
\definecolor{dialinecolor}{rgb}{0.000000, 0.000000, 0.000000}
\pgfsetstrokecolor{dialinecolor}
\node[anchor=west] at (29.616363\du,18.864369\du){\tiny $w_K^i$};
\end{tikzpicture}
\caption{Overview of the learning algorithm.}
\label{fig:em_algorithm}
\end{figure}

%% file: experiments.tex
We perform experiments in two main settings.
Most importantly, we investigate the practicality of the approach by
applying it to data from a large scale workplace experiment lasting
for a month, with most observations collected during a period of several days.
In addition, we present results from a smaller controlled experiment 
where we have annotated the positions of all movable objects in each
time step. In the following, we will describe the experiments
and how we benchmark the system using the annotations.

\subsection{Parameters}
\label{sec:parameters}

\input{parameter_table}

The parameters of the tracker are laid out in Table \ref{table:parameters}.
The parameters are the same as in previous work \cite{bore2017tracking}, except for the
spatial measurement noise $\sigma_r$, which we found to be higher than
expected during our initial learning experiments.
The intuition behind the values of the $p_\text{birth}$ and $p_\text{death}$
parameters is that they should be lower than $p_\text{jump}$.
Otherwise, it would be more likely
for a target to be inferred as dead when it jumped, only for an identical
target to be born at a new location.
$A_k$ is the support of the uniform spatial noise, and can be
estimated as the mean size of the locations in the experiments.
In our case that corresponds to about $A_k = 20 m^2$.
The feature clutter support $S^f$ and the initial value of the
feature measurement covariance $\mathbf{R}_k^{f}$, are estimated
from a dataset of annotated object instances \cite{ambrus2015unsupervised}.

\subsection{Tracker Initialization}
\label{sec:initialization}

When initializing the filter, we do not know how many objects are within each
room. We instead rely on the filter's birth mechanism to identify the targets.
But this presents a problem, as the $p_\text{birth}$ parameter needs to be tuned
high enough for all the objects present at the start to be quickly identified.
A value that is high enough for this purpose
will incorrectly lead to some measurements being
explained as new targets, instead of as jumps.
Instead, we apply a model for the number of visits
needed to observe all targets at a location.
In the experiments, we use a Poisson distribution with mean number of visits $\lambda = 1$.
By adding the probability of not having observed all targets to $p_\text{birth}$, we get a higher birth probability
for the first one or two observations.

\begin{figure*}[thpb!]
	\centering
 	\includegraphics[width=0.99\linewidth]{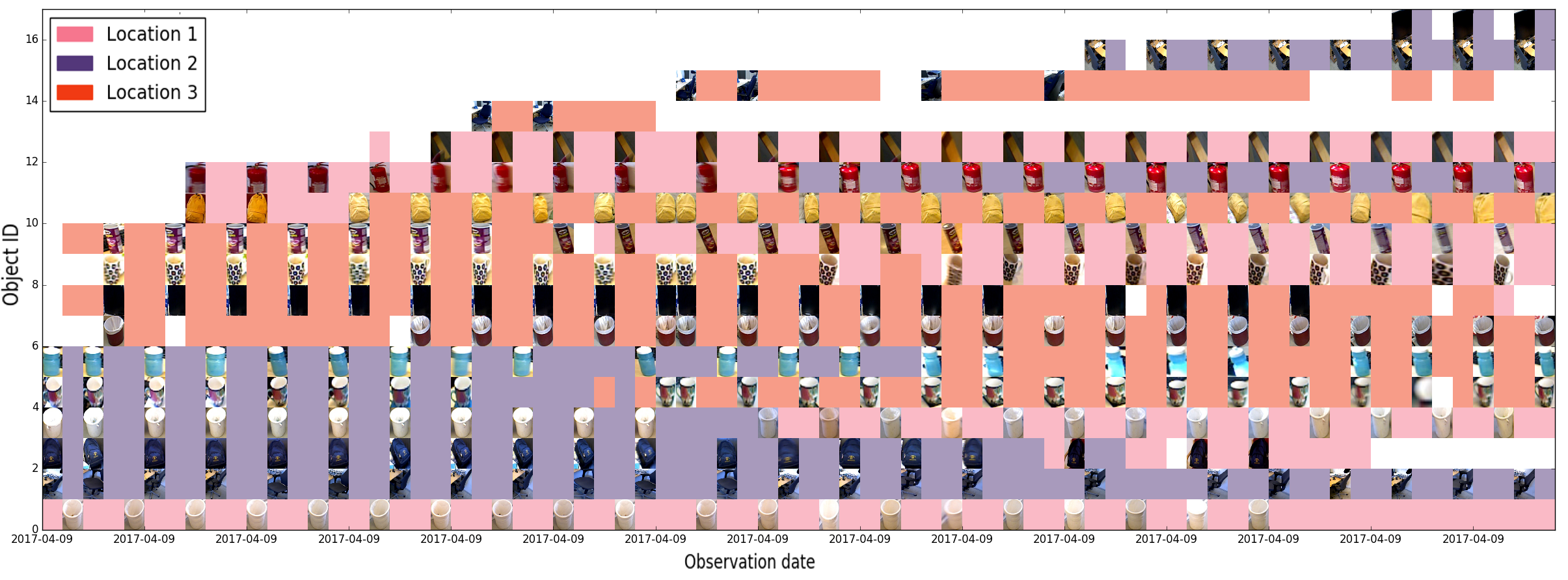} 
    \caption{Estimated associations and locations in each time step in the annotated data.
             Associations are defined as measurements that the weighted majority
             of the particles have associated with the same target.
             Each row represents an object and each color its estimated location. The images
             represent the estimated associated measurements of the objects. Note that jumps happen whenever
             an object's color shifts. All the inferred jumps in the sequence are correct.
             The main error can be seen in the uppermost row, which corresponds to the monitor that
             was initially tracked as the target in row 8.
             Rather than inferring the jump, the tracker believes that it is a new object.}
    \label{fig:varying_timestep_associations}
\end{figure*}

\subsection{Detections and Features}
\label{sec:detections}

The robot moves around the environment and visits one location
$l_k^y \in \mathcal{L}$ at each time step $k$. It observes the
location using a PTU-mounted RGBD camera and registers the frames
to form a local 3D map, see \cite{ambrus2015unsupervised} 
for details. The map is then
registered with those from time step $k-1$ and $k+1$. 
These maps can then be compared using the change detection
from \cite{ekekrantz2017segmentation} to identify objects that
have moved. Using the temporal segmentation logic
from \cite{bore2017tracking}, we then extend the
detections to any observations where the detected
objects did not move, thus allowing us to consistently detect
all objects that moved at any point in our observations.
For each time step, the detection method outputs
a set of object positions within the environment,
together with corresponding RGBD image segments.

From the RGB image segments, we extract feature vectors, one for each
detection. We feed the images through an off-the-shelf
neural network \cite{szegedy2016rethinking}, and extract the
tensor output from the last bottleneck layer. To produce a
representation suitable for tracking, we use t-SNE 
\cite{maaten2008visualizing} to reduce the 2048 dimensional network
tensor to three dimensions.
In order to learn a more stable mapping, we augment the 
feature set with additional examples from
\cite{ambrus2015unsupervised}.
As discussed in \cite{bore2017tracking},
the variation of the features of one object is well represented
by a multivariate Gaussian, motivating our use of a Kalman filter
to estimate the feature distribution. These features together with
the 2D $x$-$y$ centroids of the detected RGBD segments
make up the point measurements $\hat{\mathbf{y}}_{k, m}$.

\subsection{Performance Metric}

To benchmark the tracker, we use the \textit{Clear Multiple Object Tracking} metrics
\cite{bernardin2008evaluating} proposed by Bernardin and Stiefelhagen.
The metric builds on the association of tracked estimates to measurements
in each time step.
We use the Hungarian algorithm to find an assignment such that the sum of
distances between position estimates and measurements is minimized.
For the metric, we compute the number of false positives $N_\text{fp}$, mismatches $N_\text{mm}$,
and false negatives $N_\text{fn}$, giving us a ratio for each of these measures.
The \textit{MOTA} score is then defined as
$1 - \frac{N_\text{fp}}{N_K} - \frac{N_\text{fn}}{N_K} - \frac{N_\text{mm}}{N_K}$,
with $N_K$ being the true number of object observations in the sequence.
We also study the \textit{MOTP} score,
which is simply the mean distance between the estimated positions and
the true target positions \cite{bernardin2008evaluating}.
There is one difference in how we use the MOTA score as compared to
\cite{bernardin2008evaluating}. While they only count mismatches when the
initial errors is made, e.g. when two targets are crossing paths, we count
them through the entire sequence. This is important in our application as it
demands that we track the same target through the full sequence.
Since we are tracking a variable number of targets, there is not a unique
correspondence between annotated trajectories and filter trajectories. Again,
we find such a correspondence as that which maximizes the MOTA score.

\subsection{Baseline}

For the multi-target tracker, we have implemented a
non-probabilistic baseline based on the GATMO system
\cite{gallagher2009gatmo}. This tracker does not handle
noise, and thus associates every measurement with a
target. If there is currently no target nearby
that can explain the measurement, a new target track
is initialized. The targets may also jump; if there is no
measurement to explain a target at one location, it can
be associated with similar measurements in any position.
The system can be characterized as building
on the same mechanisms that we present here, but with
non-probabilistic spatial tracking and feature matching.
The tracker uses a threshold on the feature distance, $\tau$ to
decide if a measurement can be associated with a target, or
if a new target should be initialized. $\tau$ has been tuned
for the baseline to perform well on our annotated data set.
Based on the results in \cite{bore2017tracking}, we expect
the baseline to perform well with few clutter measurements and
discriminative features, and worse in more realistic settings.

\subsection{Controlled Experiment}

In the first experiment, we have an autonomous robot patrolling between three
locations in an office environment, all corresponding to a room.
In order to guarantee a wide variation of motion, we
have manually moved objects between the robot's visits
to the rooms. This also means that we have ground
truth knowledge of how many objects moved and when.
Within the environment, there are 15 objects,
moving around locally as well as jumping a total
of 14 times. The objects are visually diverse, but there 
are multiple trash cans, chairs and mugs in the data set.
The diversity leads to the neural network features being highly discriminative,
which we have also confirmed by inspecting the feature vectors.
Since there are few clutter measurements and the features
contain minimal noise, we expect the baseline to perform well on this dataset.

The MOTA score is computed by comparing
the filtered estimates with annotated positions.
Our aim in this
experiment is to validate that the filter can correctly infer the number of objects and and that learning improves
our performance metric on the annotated data.
We learn the parameters $\mathbf{R}^f$, $\sigma_q^s$, $p_\text{jump}$ jointly
as well as separately to decide which one has the greatest effect on
the measured performance.
In addition, we investigate if the EM algorithm converges
when initialized with different parameter values.
To plot value the value of the three-dimensional covariance $\mathbf{R}^f$, we approximate a
one-dimensional standard deviation with $\sigma_q^f = | \mathbf{R}^f |^{\frac{1}{2D}}$ (with $D=3$).

\subsection{Uncontrolled Experiment}

In the second experiment, we have a real world deployment 
of a mobile robot in an office environment.
The robot patrolled this environment and collected a total of 103
local 3D maps in 10 different locations, mostly during a period of three days.
All dynamics in the environment
are entirely due to the natural processes in this workplace.
Since we do not know which objects moved during the experiment,
we instead present qualitative results of the associated object observations.
Further, we study the influence of the learnt
parameters on the expected likelihood and the
qualitative results.
This also allows us to
investigate if learning the parameters for an individual
environment helps improve performance.

%% file: parameter_table.tex
\definecolor{lightgray}{gray}{0.9}


\begin{table}[htpb]
\begin{center}
\rowcolors{1}{}{lightgray}
\begin{tabular}{r|rrrrrrrrr}
 Param & $p_{\text{jump}}^*$ & $p_{\text{meas}}$ & $p_{\text{birth}}$ & $p_{\text{death}}$ & $\sigma_q^*$ & $\sigma_r$ & $A_k$ & $\mathbf{R}_k^{f*}$, $S^f$ \\
  \hline
  Value & 0.03 & 0.98 & 0.01 & 0.005 & 0.15 & 0.35 & 20 & estimated \\
\end{tabular}
\end{center}
\caption{The parameters used in the experiments. \footnotesize{\textsuperscript{*}These parameters are 
initialized with these values, but they are then learnt in the EM scheme.}}
\label{table:parameters}
\end{table}

%% file: results.tex
We initially planned on evaluating the learning of parameters
of individual instances. However, when applied on the presented
data sets, the variance of the results turned out to be too
large to draw any meaningful conclusions.
For example, in the real-world experiment, we only have around five
observations for many of the tracked objects, which is
insufficient for reliable covariance estimation.
In these results, we instead focus on learning one
set of parameters for all object instances in the environment.
This allows for a more detailed analysis on the influence of dynamics learning.
Nonetheless, the results presented here also indicate what
to expect with equivalent amounts of data for individual objects.

Unless otherwise stated, the experiments are performed
ten times, and statistics from those runs are presented.
In the graphs, we present the mean values of the runs,
and the area corresponding to one standard deviation from the mean.

\begin{figure}[h!]
	\centering
 	\includegraphics[trim=40 0 60 20,clip,width=0.99\linewidth]{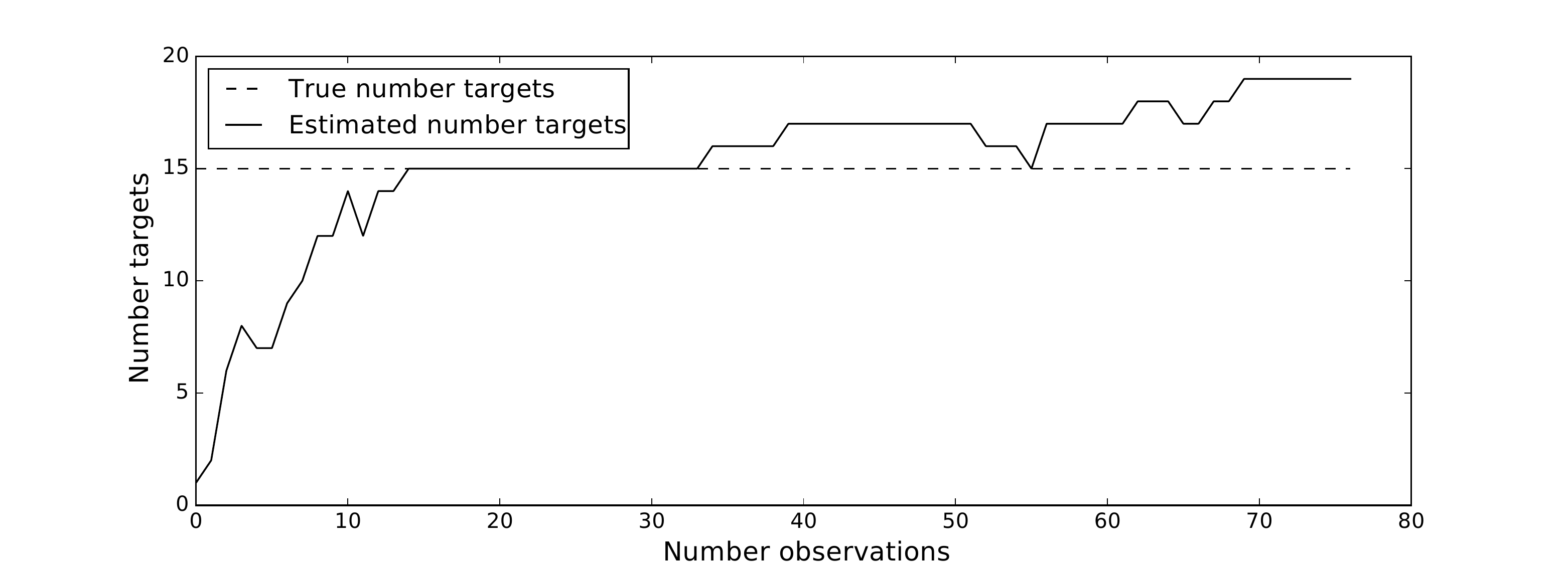} 
    \caption{Typical (one run) estimated and true number of targets in annotated data.}
    \label{fig:estimated_nbr_targets}
\end{figure}

\subsection{Controlled Experiment}

\subsubsection{Estimating number of targets}

To begin with, we investigate if the filter correctly infers the
number of objects in the scene. This may not be the case, as the
filter may interpret some measurements as old objects
jumping around rather than new ones appearing.
While we observed that the behavior is
somewhat dependent on the birth rate $p_\text{birth}$, we saw that
a value lower than $1-p_\text{meas}$, around $0.01$, yields good results.
In Figure \ref{fig:estimated_nbr_targets},
we see a typical progression of the estimated number of targets over time.
Since all targets are present from the beginning, a perfect result would
be for the filter to immediately estimate that there are 15 targets present.
Such a filter would be too eager,
since it would require all measurements,
including the clutter, to be classified as targets.
Instead, we see that our filter slowly stabilizes at
one or two targets from the correct value.
It converges around time step 25, which corresponds
to approximately 8 observations of each target.
In Figure \ref{fig:varying_timestep_associations}, we see
the estimated existing targets and their inferred associations over time.
Here, we see the filter identifies ten jumps correctly. It
incorrectly infers only one new target birth instead of a jump.
Together, the results demonstrate that we can infer the number
of targets at the same time as tracking their jumps.

\begin{figure}[h!]
	\centering
 	\includegraphics[trim=50 0 70 20,clip,width=0.89\linewidth]{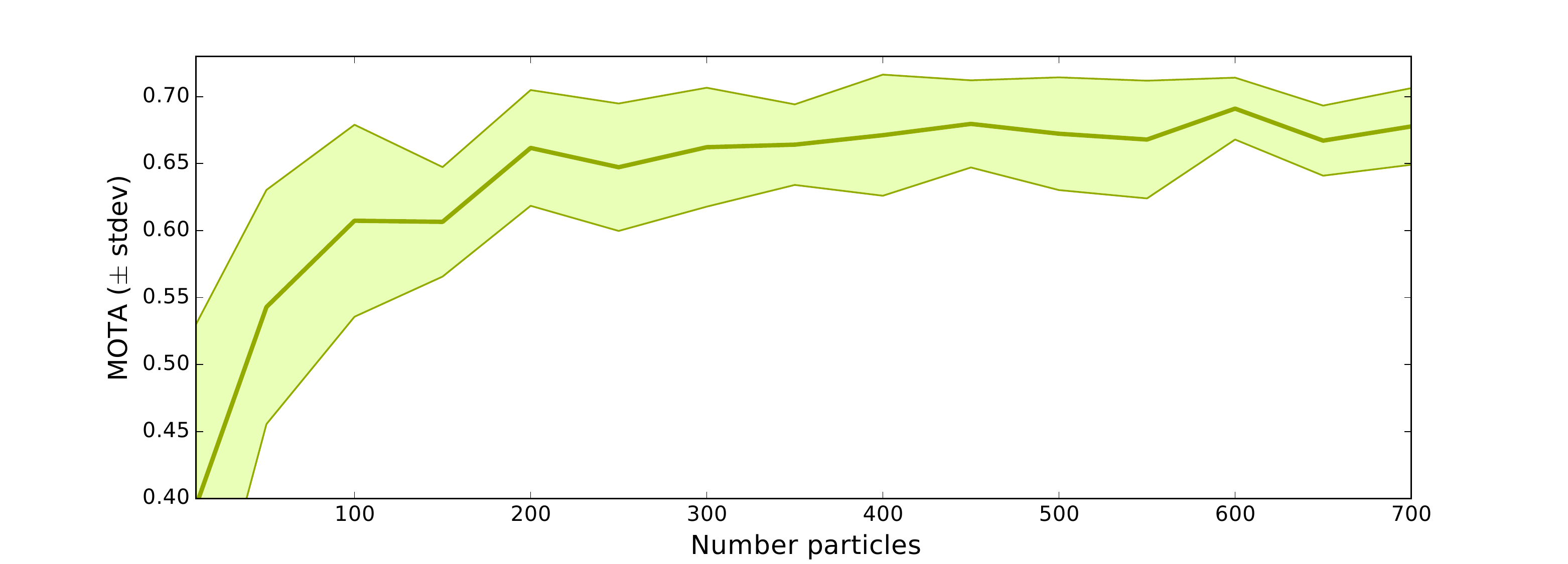} 
    \caption{The MOTA score as a function of the number of particles. It stabilizes
    with around 300 particles.}
    \label{fig:number_particles}
\end{figure}

\subsubsection{Number of particles}

\begin{figure*}[thpb!]
\centering
\begin{minipage}[t]{.49\linewidth}
    \centering
 	\includegraphics[trim=0 0 0 20,clip,width=0.99\linewidth]{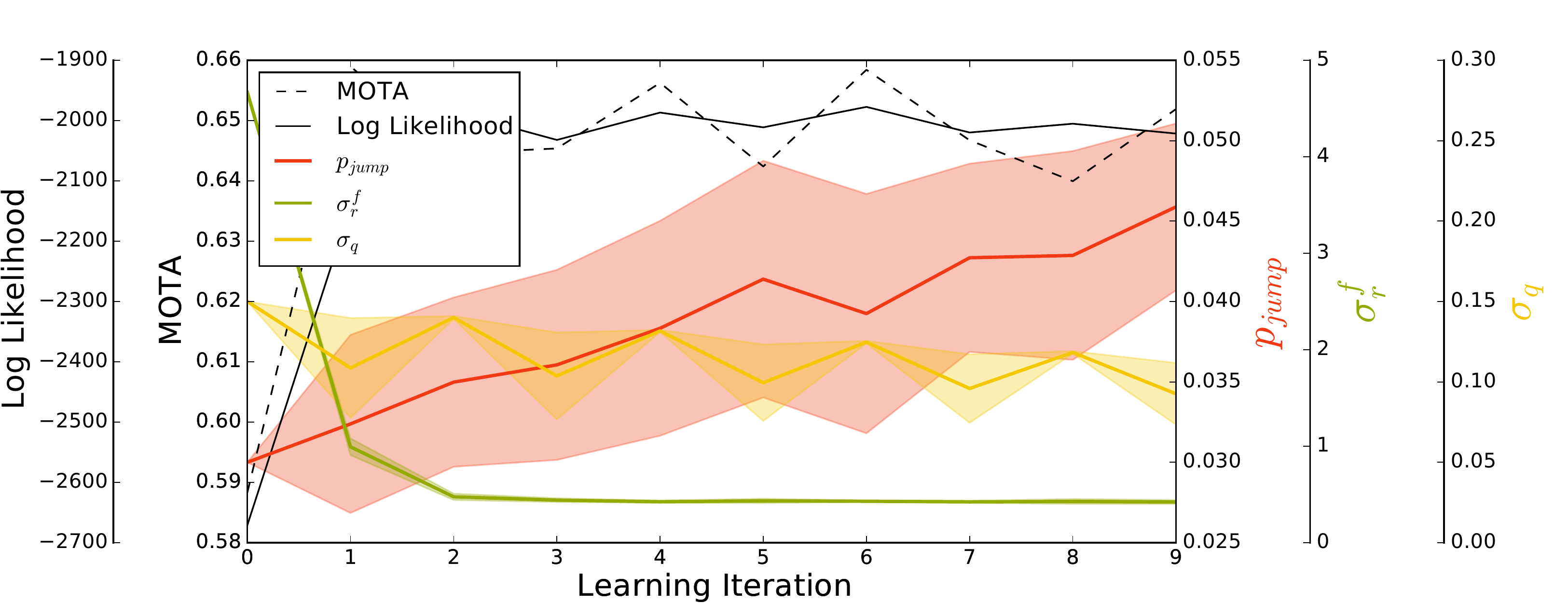} 
    \caption{The iterations of the EM
    		 algorithm in the controlled environment.
    		 The MOTA score changes with each
    		 iteration, as well as the value of the parameters.
    		 Note that the $\sigma_q^f$ parameter converges after about three steps,
    		 while $p_\text{jump}$ takes longer.
    		 Both the estimates and the MOTA score seem to have stabilized after about 6 iterations.}
    \label{fig:learning_all}
\end{minipage}\hfill
\begin{minipage}[t]{.49\linewidth}
    \centering
 	\includegraphics[trim=0 0 150 20,clip,width=0.83\linewidth]{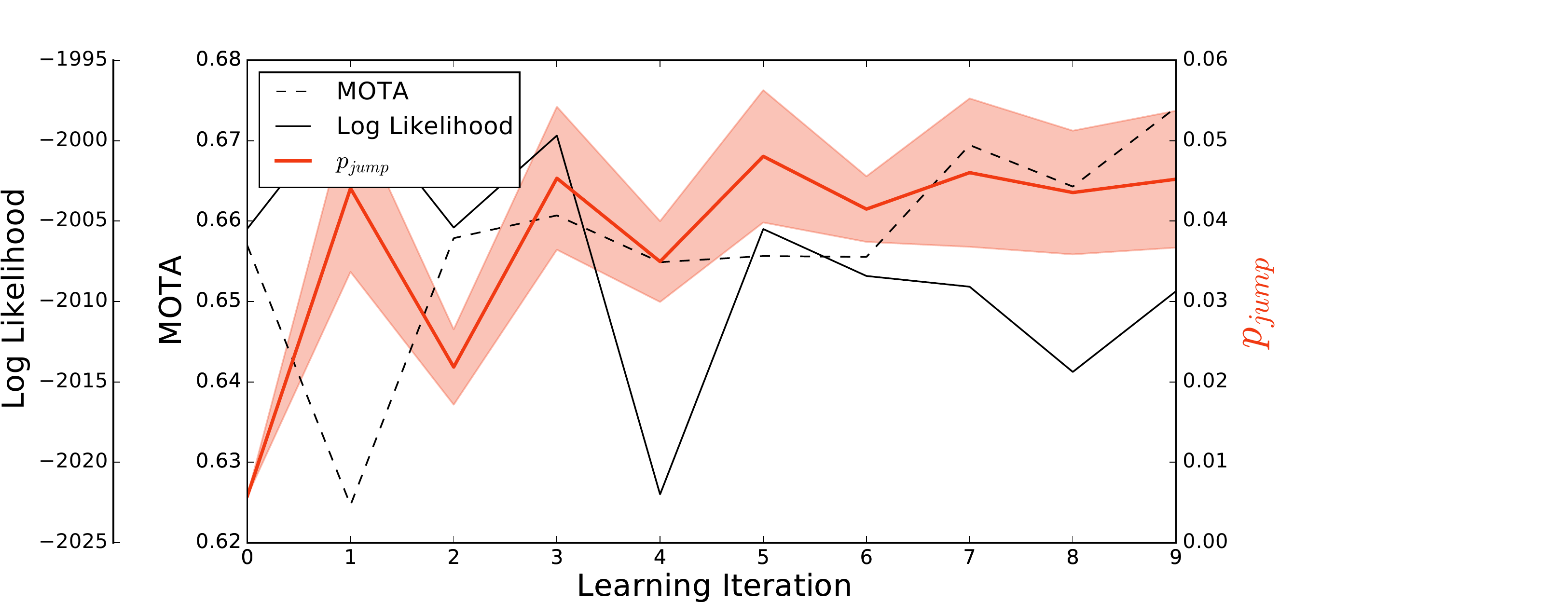}
    \caption{The EM algorithm on controlled data, applied only on the jump parameter
    	     with a poor initial value of $p_\text{jump} = 0.005$.
    	     We use the learnt $\mathbf{R}^f$ estimate in this experiment.
    	     To reduce noise, the trend is averaged over 50 experiments.
    	     The MOTA score increases, but the expected likelihood is noisy,
    	     and is not indicative of the filter performance. }
    \label{fig:learning_pjump}
\end{minipage}
\end{figure*}

We also investigated the influence of the particle set size on the
MOTA score, as well as the effect of the number of Gibbs iterations.
In Figure \ref{fig:number_particles}, we can see that increasing the
number of particles initially has a significant effect on performance,
and that it stabilizes around 300 particles.
If we compare with the results from \cite{bore2017tracking}, where
the number of targets is fixed, adding variable number of targets via
sampling of the birth and death process seems to increase the need
for sufficiently many particles. While the measure stabilizes
around 300 particles in both cases, the score is comparatively much lower
with fewer particles with the new general method. The effect of the
number of Gibbs iterations is more marginal. We did not see any
measurable effect when comparing numbers in ranges between 50 and 200
Gibbs iterations. In the controlled experiment, we have therefore settled on
50 iterations and 300 particles. In the larger
real world experiment, we instead use 100 particles in order to run
the learning iterations in a reasonable time.

\subsubsection{Learning parameters}

\input{learning_parameters_table}

The learning of parameters need to be carefully dissected, as changing the
value of one parameter can affect the optimal value of the other parameters.
It is also of interest to investigate which parameters have the largest
effect on our performance measure. In Figure \ref{fig:learning_all}, we
see how the values of all three parameters vary when learnt jointly.
We also see how the likelihood and MOTA score increase with the number
of EM iterations, with the MOTA starting at $0.59$ and improving to $0.65$.
After about five iterations, all parameter values seem to have
approximately converged, with the change in feature measurement covariance
being the most significant, as it decreases by an order of magnitude.
In general, we have seen that learning of $\mathbf{R}^f$
contributes to the largest increase in the MOTA score with these initial values.
While the learnt $\mathbf{R}^f$ value is significantly lower than the initial value,
it does not seem to underestimate
the variance, as the MOTA score continuously increases.

\begin{figure}[h!]
	\centering
 	\includegraphics[trim=20 10 0 30,clip,width=0.99\linewidth]{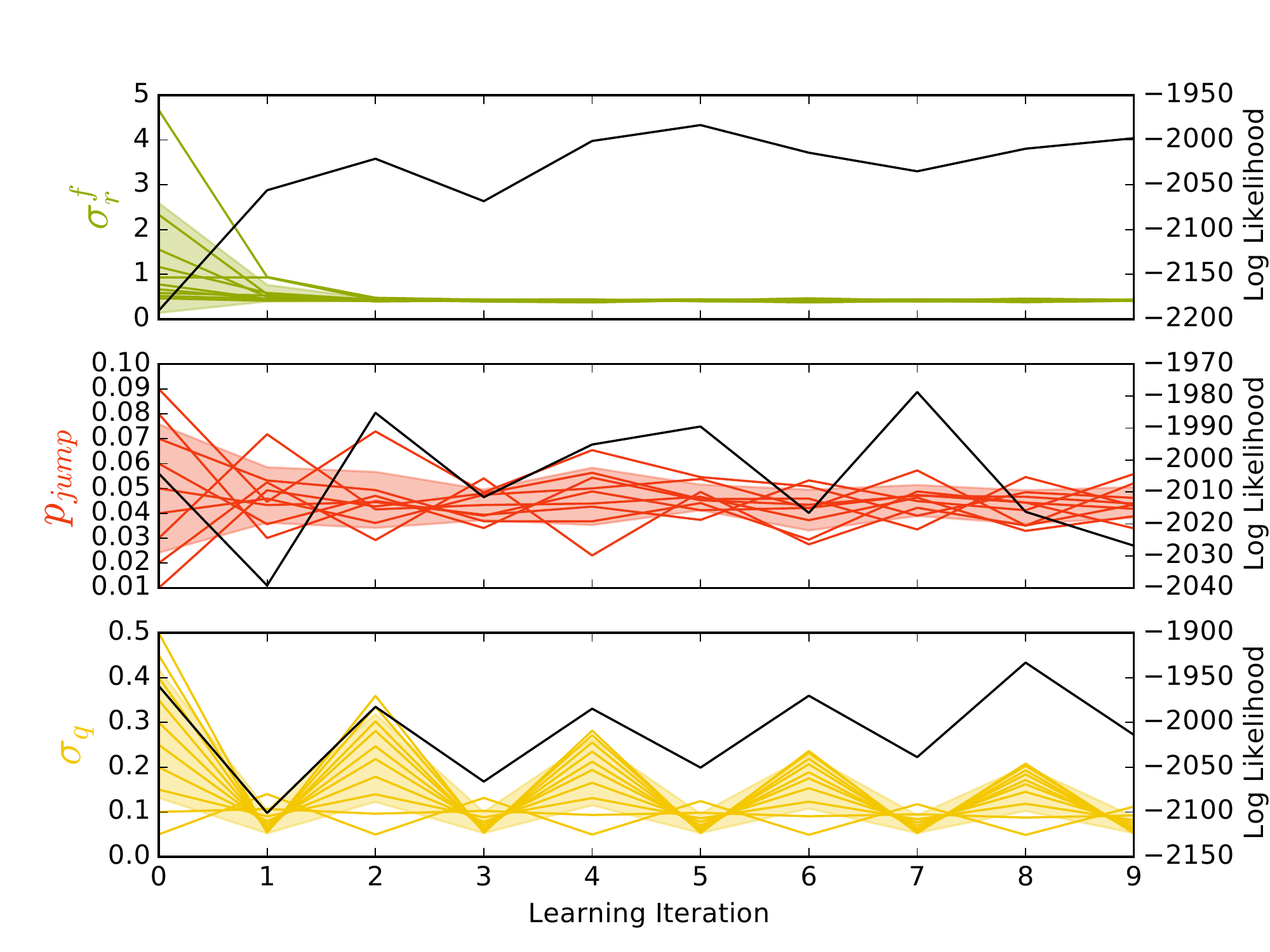} 
    \caption{Learning of the three parameters separately, on the
             controlled data. We see that even with a wide spread
             of initial values, the parameter estimates converge
             to similar values. The expected likelihood (black) increases except
             for $p_\text{jump}$, where the variance seems to cancel out any trend.}
    \label{fig:various_learning_separate}
\end{figure}

\begin{figure*}[thpb!]
	\centering
 	\includegraphics[width=0.99\linewidth]{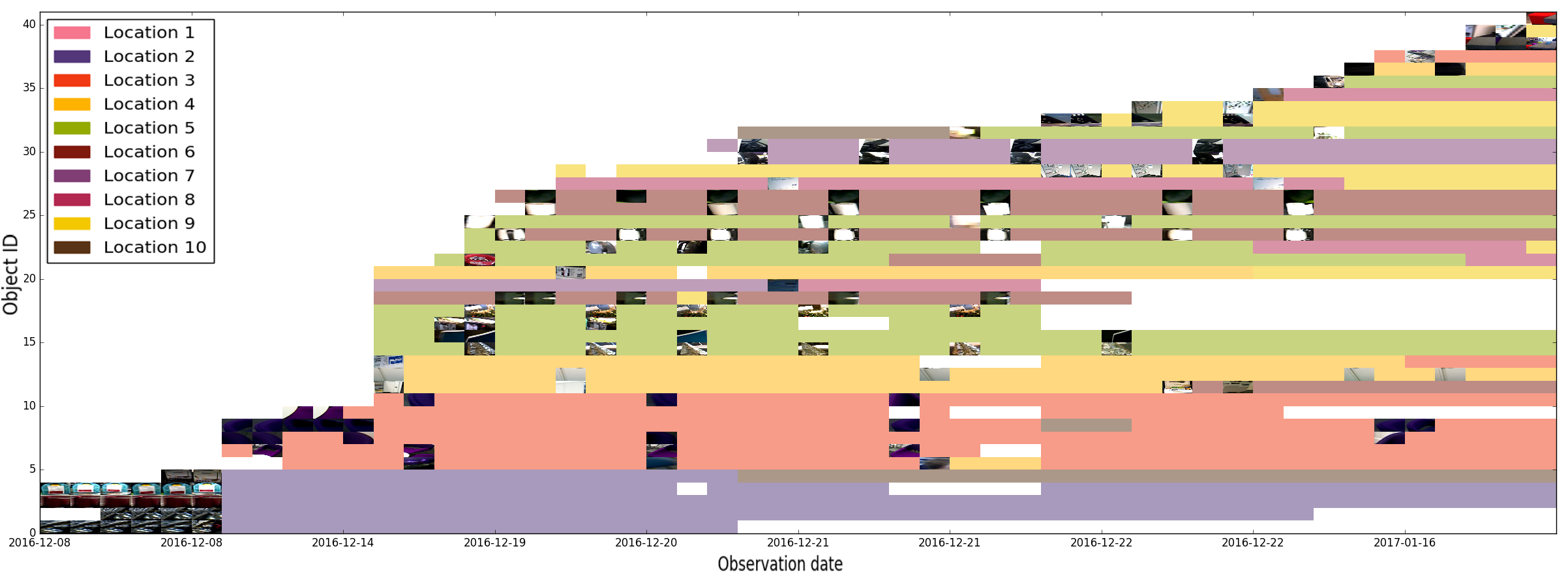} 
    \caption{The estimated associations and locations in each time step, after ten iterations of parameter learning.
             The data is from the long-term deployment in an office environment, and the observation 
             frequency of each location is low. Note that the method correctly infers that most objects
             stay in the same location. In Figure \ref{fig:tsc_object_images} below, we show larger images
             of some of the associated measurements.}
    \label{fig:tsc_timestep_associations}
\end{figure*}

\input{tsc_object_figure}

In Table \ref{table:learning_parameters}, we see more variations
of learning different parameters, with the parameters initialized
with the values in Table \ref{table:parameters}, and learnt
separately or jointly. When learnt separately, the initial values for $p_\text{jump}$
and $\sigma_q$ were close to the learnt estimates, with $p_\text{jump}$
going from $0.30$ to $0.44$ and $\sigma_q$ from $0.15$
to $0.08$. This is reflected in the table by the learnt parameters
achieving a MOTA score similar to the initial parameters.
As mentioned, learning of $\mathbf{R}^f$ results in the largest difference,
and the learnt parameter contributes to a significant increase in the
MOTA score, from $0.60$ to $0.65$. When learning all the parameters
jointly, we achieve a similar score, indicating that the learning
of $\mathbf{R}^f$ contributed to the largest improvement.
In Figure \ref{fig:learning_pjump}, we see that learning
$p_\text{jump}$ by itself from a highly unrealistic
initial value also results in an improved MOTA score.
Moreover, in Figure \ref{fig:various_learning_separate}, we see that
all parameters converge to a single estimate, even when the initial
values are spread across a range. The result shows that we can expect
a performance similar to the ones presented in Table \ref{table:learning_parameters},
even with poor initial values. This removes the need
to manually tune parameters, as the system can adapt to the data at hand.

In general, the expected log likelihood seems to continuously
increase when learning $\sigma_q$ and $\mathbf{R}^f$,
see Figure \ref{fig:various_learning_separate}. In the case
of $p_\text{jump}$, we see that it varies more. This is due to
the complexity of the experiment setup, where changing the $p_\text{jump}$
parameter can have unforeseen consequences. These may occurr for example if the jump
rate is not homogenouous throughout the sequence, causing a globally
correct rate to be suboptimal in some parts. The results in
Figure \ref{fig:learning_pjump} show that the MOTA score as
a function of $p_\text{jump}$ may have several local maxima.
We also observe that the change in expected likelihood does
not always seem to be indicative of the MOTA score in the
case of $p_\text{jump}$, and that it may improve even as the
likelihood decreases.

The baseline achieves the best score on this dataset (see Table \ref{table:learning_parameters}). The reason for this is two-fold.
First, there is minimal visual ambiguity between the objects in this dataset,
reducing the need for probabilistic data association.
Secondly, the baseline estimates all measurements to be targets directly,
with no filtering to ensure that the measurements originate from persistent objects.
In contrast, the proposed tracker typically requires about three
observations before a new object is inferred. We can see
this in Table \ref{table:learning_parameters} through that the
baseline has a higher mismatch rate, while the others have
a higher miss rate. This indicates that the baseline performed
better in the first part of the sequence when the probabilistic
tracker is still unsure, therefore failing to associate the measurements
with any object. In the later part, the baseline is instead
more likely to associate measurements with the wrong object, leading
to the observed higher mismatch rate.
With a longer sequence, the probabilistic method would therefore
likely perform better than the baseline. The association threshold
$\tau$ was also tuned for this benchmark, while the probabilistic
method can automatically adapt to the variance in the data.

\subsection{Uncontrolled Experiment}

In the uncontrolled experiment, the conditions differ from the first experiment.
Primarily, we have more than three times as many locations,
which means that each location is
visited with a lower frequency. In addition, several locations are visited
even more seldom, with some being visited only once. All of the
movement is completely natural and happens over a period of several days.
In Figure \ref{fig:tsc_timestep_associations}, we see the inferred
locations and associations after ten iterations of EM.
This time, there are longer periods between associations of measurements to one object.
Looking at the bottom rows, we see that at least one of the locations is only
visited once at the beginning. Nevertheless, the tracker is able to track
several of the objects throughout most of the period, which stretches over
Christmas. Indeed, rows 17 and 14 correspond to a decorative reindeer and
Christmas tree respectively (see larger pictures in Figure \ref{fig:tsc_object_images}).
Starting from January, they seem to be absent
from the environment and the filter finds no more associations.
It terminates the reindeer track but believes that the Christmas tree is still there.
Moreover, in Figure \ref{fig:tsc_object_images} we see that two paper rolls
are included among the tracked objects. Both of them stand on kitchen counters,
and people use them and put them back during the period. This demonstrates that the
method can track objects also when their positions vary.
Locations 5 and 6 correspond to the two kitchens.
They are visited more frequently than the other locations, and this
seems to result in more object estimates in those positions. Drawing from
this result and those in the previous section, the filter generally
behaves better when the observations are more equally distributed among the rooms.
The results demonstrate that the method can be applied to real
workplace data, and that the filter can track objects over periods of at least several days.

\begin{figure*}[thpb!]
	\centering
 	\includegraphics[width=0.99\linewidth]{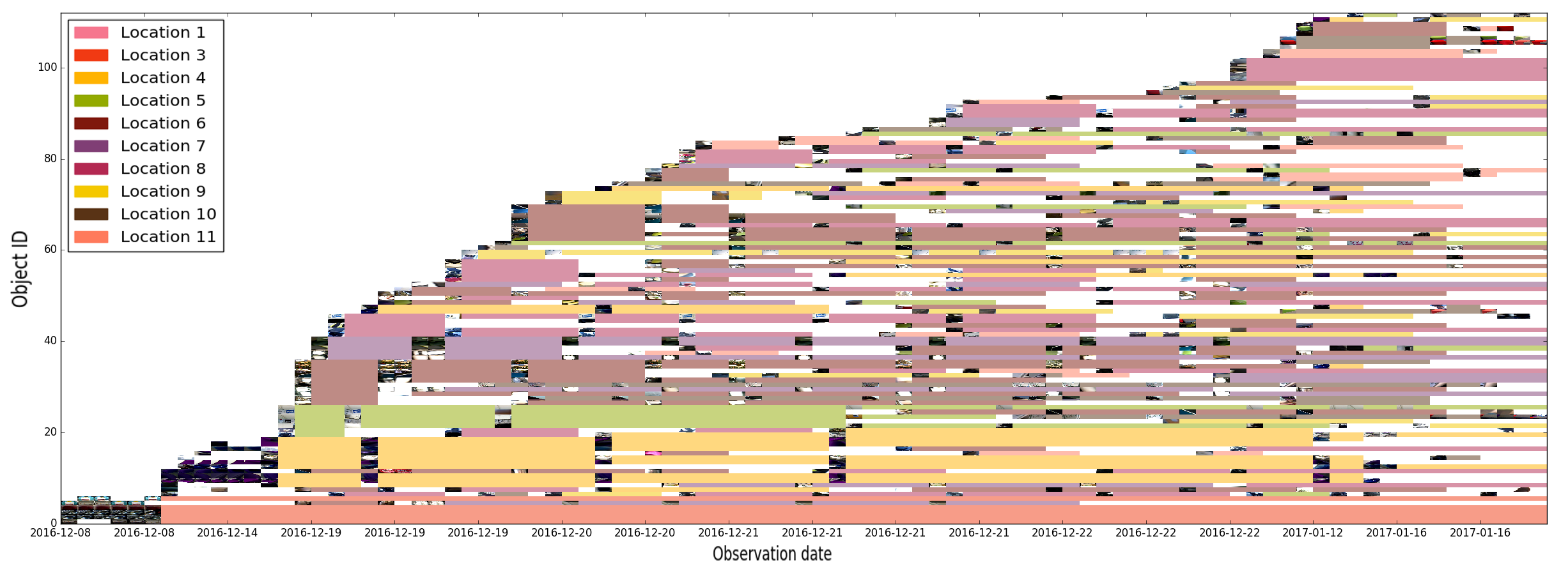} 
    \caption{The target associations of the baseline tracker on the
 	         real-world workplace dataset. The association images
             are hard to make out. While several targets are tracked
             consistently, the tracker fails to deal with spurious
             object measurements, and initializes a wealth of new   
             targets.}
    \label{fig:tsc_baseline_associations}
\end{figure*}

The associations of the baseline when applied to the uncontrolled
dataset are illustrated in Figure \ref{fig:tsc_baseline_associations}.
Although, the images are hard to make out, we see that the baseline
initializes more than twice the number of targets as the proposed method.
Several of the targets only seem to be present for a short amount of
time, indicating that they correspond to non-persistent objects.
Looking at the actual associations, few of the targets are consistently
associated with only one object across the entire sequence,
as opposed to the probabilistic tracker. The results indicate that
the baseline incorporates more false positives and mismatches.
This is due to its inability to handle clutter, and to adapt to the
different noise characteristics of this data set.

\begin{figure}[thpb!]
	\centering
 	\includegraphics[trim=40 0 0 20,clip,width=0.99\linewidth]{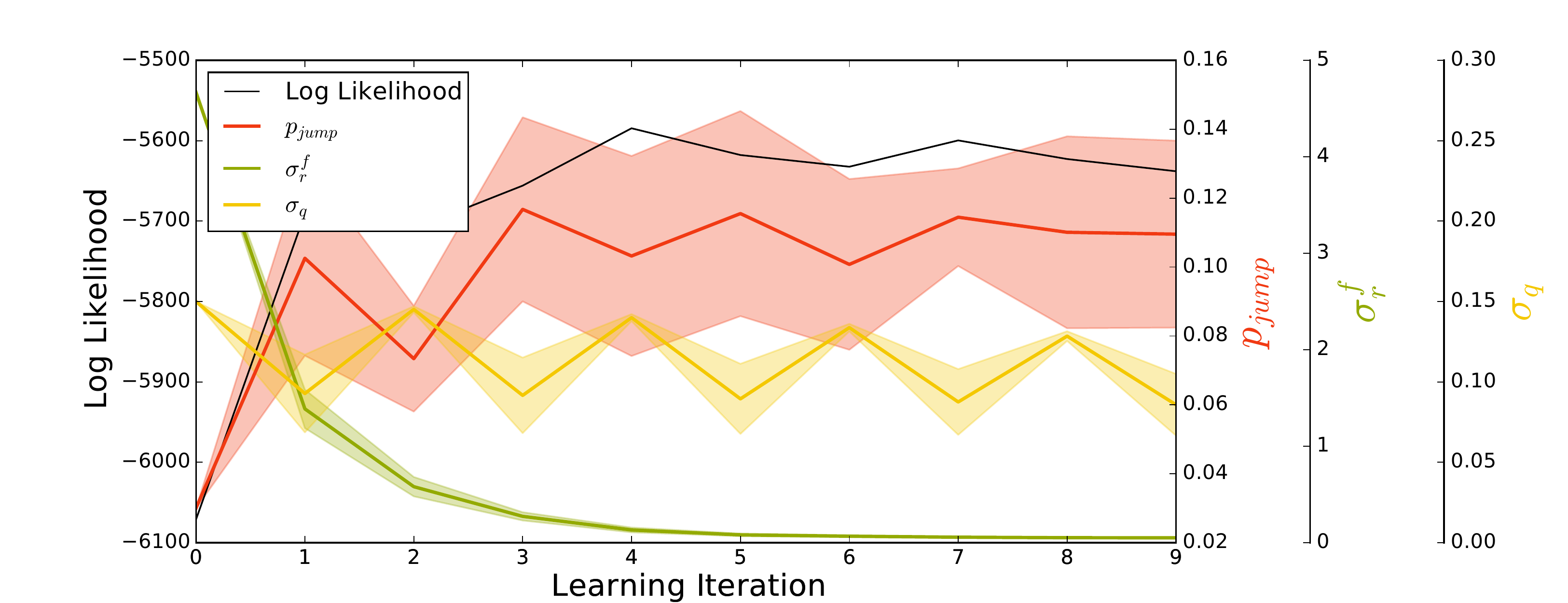} 
    \caption{Parameter estimates when the EM algorithm is applied
             to the uncontrolled data. Since we have no annotations here, we
             can present the likelihood, but not the MOTA score.
             We see that all parameters converge after a few iterations.
    		 Again, the feature measurement covariance was clearly overestimated,
    		 and is even lower in this experiment, at $\sigma_r^f \approx 0.05$.
    		 }
    \label{fig:tsc_learning_all}
\end{figure}

We also analyze the behavior of the learning scheme on the uncontrolled data. Since we
have no annotations, we will instead look at the convergence
of the estimated parameter values and how they affect the qualitative results.
If we look at results produced using the initial
parameter values, they look similar to the ones in Figure \ref{fig:tsc_timestep_associations}.
The main difference lies in that a few targets have been associated
with two different objects.
Looking at learning iterations in Figure \ref{fig:tsc_learning_all}, we see that
the values develop similarly over time as to the controlled experiment.
Again, the largest difference occurs in the measurement covariance,
with $\mathbf{R}^f$ being another order of magnitude smaller this time.
This explains the mismatch of some of the observations with the initial parameter values.
While $p_{jump}$ is larger than in the previous experiment, one has
to factor in the $\tilde \delta$ adjustment from Section \ref{sec:estimation}.
The mean estimated time away from a location
was estimated to be $\tilde \delta \approx 3$ in the 
controlled experiment, and $\tilde \delta \approx 7$ in this experiment.
With that in mind, the estimated jump rate is about comparable between
the two experiments, and corresponds to about $1.7$ jumps in average
during this period.

\begin{figure}[thpb!]
	\centering
 	\includegraphics[trim=20 10 0 30,clip,width=0.99\linewidth]{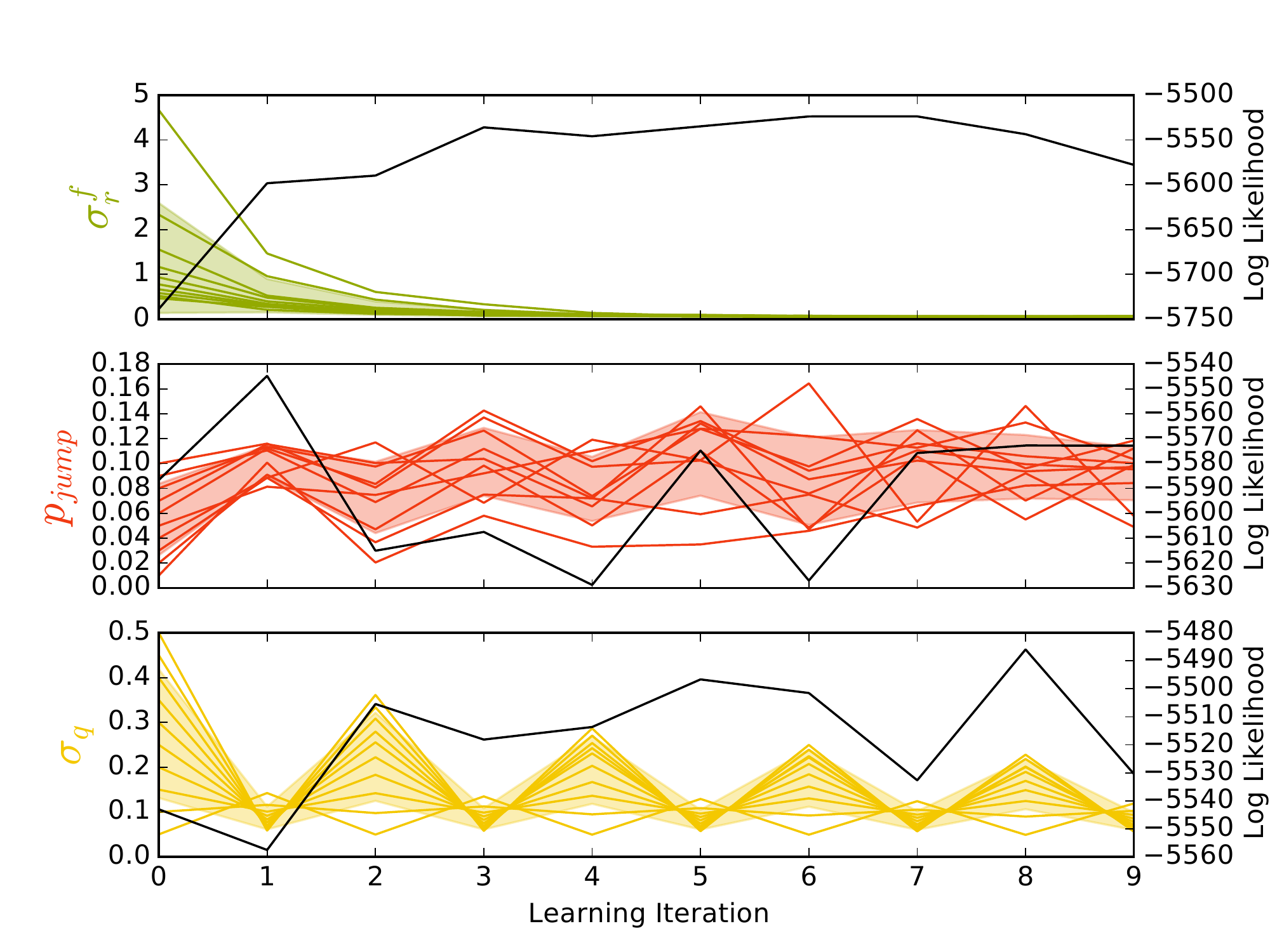} 
    \caption{Learning the parameters from a range of initial
             values on the uncontrolled experiment. The parameters all converge to some range.
             The variance of the $p_\text{jump}$ estimates is higher than in the
             controlled experiment. It is amplified by the $\tilde \delta$ factor.}
    \label{fig:tsc_learning_separate}
\end{figure}

In Figure \ref{fig:tsc_learning_separate}, we again see that the EM algorithm
converges from different initial values. In this case, since the initial values
of $\mathbf{R}^f$ is off by orders of magnitude, the filter has to rely on
the spatial tracking to a high degree during the first EM iterations.
This showcases a useful synergy, where our integration of spatial and visual
models can work in tandem to reinforce each other. Once the feature covariance
has adjusted, it is required to estimate an accurate value for $p_\text{jump}$,
as jumps are inferred largely through visual similarity.
In conclusion, we see that the parameters all stabilize and that
the expected likelihood increases for $\sigma_q$ and $\mathbf{R}^f$.
The qualitative results further confirm that the learnt dynamics
improve our modeling of the environment.

%% file: learning_parameters_table.tex
\definecolor{lightgray}{gray}{0.9}

\begin{table}[htpb]
\begin{center}
\rowcolors{1}{}{lightgray}
\begin{tabular}{r|rrrrr}
 System & MOTP\cite{bernardin2008evaluating} & Miss rate & False pos. & Mism. & MOTA\cite{bernardin2008evaluating} \\
  \hline
  Initialization & 0.14 & 0.25 & 0.02 & 0.13 & 0.60 \\
  Learn $p_\text{jump}$ & 0.15 & 0.25 & 0.03 & 0.13 & 0.59 \\
  Learn $\sigma_q$ & 0.15 & 0.24 & 0.03 & 0.13 & 0.60 \\
  Learn $\mathbf{R}^f$ & 0.13 & 0.21 & 0.04 & 0.11 & 0.65 \\
  Learn all & 0.14 & 0.21 & 0.03 & 0.10 & 0.65 \\
  Baseline & 0.11 & 0.09 & 0.04 & 0.18 & \bf{0.70} \\
\end{tabular}
\end{center}
\caption{Comparison with the baseline and when learning none, one or all of the parameters.
         Since each learning iteration takes around 40 min, we instead benchmark the tracker with the
         learnt parameters from Figure \ref{fig:various_learning_separate} for $p_\text{jump}$,
         $\sigma_q$, $\mathbf{R}^f$, and from Figure \ref{fig:learning_all} for the joint
         learning. We run 50 benchmarks for each parameter set, and present the averages.
         $\mathbf{R}^f$ has the largest impact on the MOTA score.
         }
\label{table:learning_parameters}
\end{table}

%% file: tsc_object_figure.tex
\begin{figure*}[t!]
    \centering
    \begin{subfigure}[b]{0.095\textwidth}
        \centering
        \includegraphics[height=0.6in]{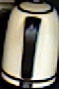}
        \caption{$l$=6,$j$=23\\Water boiler}
    \end{subfigure}
	\begin{subfigure}[b]{0.095\textwidth}
        \centering
        \includegraphics[height=0.6in]{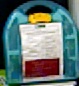}
        \caption{$l$=2,$j$=3\\Heart starter}
    \end{subfigure}
    \begin{subfigure}[b]{0.095\textwidth}
        \centering
        \includegraphics[height=0.6in]{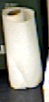}
        \caption{$l$=6,$j$=25\\Paper towel}
    \end{subfigure}
    \begin{subfigure}[b]{0.095\textwidth}
        \centering
        \includegraphics[height=0.6in]{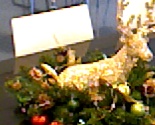}
        \caption{$l$=5,$j$=17\\Christm. deer}
    \end{subfigure}
    \begin{subfigure}[b]{0.095\textwidth}
        \centering
        \includegraphics[height=0.6in]{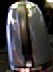}
        \caption{$l$=5,$j$=22\\Water boiler}
    \end{subfigure}
    \begin{subfigure}[b]{0.095\textwidth}
        \centering
        \includegraphics[height=0.6in]{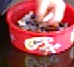}
        \caption{$l$=5,$j$=21\\Bowl}
    \end{subfigure}
    \begin{subfigure}[b]{0.095\textwidth}
        \centering
        \includegraphics[height=0.6in]{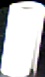}
        \caption{$l$=5,$j$=24\\Paper towel}
    \end{subfigure}
    \begin{subfigure}[b]{0.095\textwidth}
        \centering
        \includegraphics[height=0.6in]{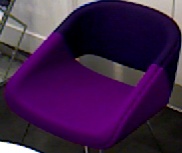}
        \caption{$l$=1,$j$=6\\Chair}
    \end{subfigure}
    \begin{subfigure}[b]{0.095\textwidth}
        \centering
        \includegraphics[height=0.6in]{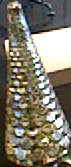}
        \caption{$l$=5,$j$=14\\Christm. tree}
    \end{subfigure}
    \begin{subfigure}[b]{0.095\textwidth}
        \centering
        \includegraphics[height=0.6in]{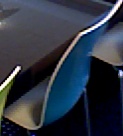}
        \caption{$l$=5,$j$=15\\Chair}
    \end{subfigure}
    \caption{Some of the discovered objects present in Figure \ref{fig:tsc_timestep_associations}.
    The object identities $j$ correspond to the the rows and $l$ are the locations.}
    \label{fig:tsc_object_images}
\end{figure*}

%% file: discussion.tex
With these results, we have shown that our filter can
estimate the number of targets at the same time as tracking
discontinuous jumps. This is an important generalization
from our previous work \cite{bore2017tracking},
and allows our filter to be applied directly on
detections from a mobile robot, with no further input.
In our case, the only inputs to our system are RGBD frames and relative
positions and the outputs are posteriors over object
positions and their discrete locations. In fact, this
information is generally available on most robots equipped
with an RGBD camera. The main adaptation that could be
necessary in some of these systems is to divide the environment
into discrete locations that can only be observed one at a time.
As we believe that rooms in indoor environments is a very
natural division, we think that this is a minor adaptation.

In the learning experiments, we saw that the EM iteration
converges on stable parameter values, with an increase
in expected likelihood and on benchmark scores. With this data,
we concluded that there were enough data points to learn
overall dynamics for all objects. If we had more data,
or more observations of one object, the algorithm
would be able to estimate parameters for individual objects.
If there is any difference in how the object
moves, this will lead to improved performance. Indeed, there
are reasons to believe that the movement of different objects
vary significantly. For instance, a door or a chair will frequently
move back and forth whenever used by a human. By contrast,
a computer monitor might be adjusted frequently, but the
movement is more slight. As more robots are deployed continuously
in large environment, we are optimistic that this kind of data
will become readily available. Our system should allow for
fine grained learning of dynamics whenever that happens.

With more data comes also the possibility of estimating how
the objects interact with the environment and with themselves.
This is an interesting venue for future work.
For example, we could imagine that there is a drying rack with
cups in the kitchen. People fetch a cup of coffee and bring it
to their offices. When finished, they wash the cups and put them
back in the drying rack. This is something that we could encompass
in our model by making $p_\text{jump}$ dependent on the origin
and destination room. In this example, if a cup is missing from
an office, it would be much more likely to jump to the kitchen
than any of the other rooms. Note that this amounts to estimating
an ordinary Markov chain, and can be easily integrated into
our EM iteration, similar to estimation of the global $p_\text{jump}$ value.

One fundamental aspect that we have left out in this treatment
is that the the parameters $p_\text{jump}$ and $\sigma_q$ should
depend on the actual time interval between $k$ and $k+1$.
If more time has elapsed, we should be more uncertain of
the object positions. We have found that our system performs
well even without this dependence, which is partly due to the
intervals mostly having similar lengths.
We have also left out parameter priors in this treatment, to demonstrate that our
algorithm converges to reasonable values even without strong priors.
However, when learning the parameters in a real robot deployment, incorporating priors
for the parameters will likely be necessary. Strong priors will enable
learning even when there are few data points, such as when
estimating models for individual objects. For example, one could
incorporate a beta prior in the estimation of $p_\text{jump}$
if there is some idea of the actual value.

%% file: conclusion.tex
In conclusion, we have presented a system for tracking general
movable objects in observations made by a mobile robot as it is
moving around a large environment. Our principled probabilistic
tracking framework produces an estimate of the number of
movable objects present as well as posteriors over the object
positions. Our results demonstrate that it is possible to track
a variable number of objects at the same time as inferring
discontinuous jumps that happen when the robot is not present.
We can minimize the need for parameter tuning
by employing unsupervised model learning. In effect, this
enables us to learn the dynamics of a particular robot environment.
It is our belief that with more robot deployments, and more data,
this system will enable learning of fine grained object dynamics.
In particular, the proposed system is capable of learning movement models
that take into account how individual objects move
with respect to their environment.
Use cases include applications such as robots for general cleaning and
fetching, or object surveillance scenarios.

%% file: ms.bbl
\begin{thebibliography}{10}

\bibitem{bore2017tracking}
N.~Bore, J.~Ekekrantz, P.~Jensfelt, and J.~Folkesson, ``Detection and tracking
  of general movable objects in large 3d maps,'' {\em arXiv preprint
  arXiv:1712.08409}, 2017.

\bibitem{burgard1999experiences}
W.~Burgard, A.~Cremers, D.~Fox, D.~H{\"a}hnel, G.~Lakemeyer, D.~Schulz,
  W.~Steiner, and S.~Thrun, ``Experiences with an interactive museum tour-guide
  robot,'' {\em Artificial intelligence}, vol.~114, no.~1-2, pp.~3--55, 1999.

\bibitem{triebel2016spencer}
R.~Triebel, K.~Arras, R.~Alami, L.~Beyer, S.~Breuers, R.~Chatila, M.~Chetouani,
  D.~Cremers, V.~Evers, M.~Fiore, {\em et~al.}, ``Spencer: A socially aware
  service robot for passenger guidance and help in busy airports,'' in {\em
  Field and Service Robotics}, pp.~607--622, Springer, 2016.

\bibitem{veloso2012cobots}
M.~Veloso, J.~Biswas, B.~Coltin, S.~Rosenthal, T.~Kollar, C.~Mericli,
  M.~Samadi, S.~Brandão, and R.~Ventura, ``Cobots: Collaborative robots
  servicing multi-floor buildings,'' in {\em 2012 IEEE/RSJ International
  Conference on Intelligent Robots and Systems}, pp.~5446--5447, Oct 2012.

\bibitem{hawes2016strands}
N.~Hawes, C.~Burbridge, F.~Jovan, L.~Kunze, B.~Lacerda, L.~Mudrova, J.~Young,
  J.~Wyatt, D.~Hebesberger, T.~Kortner, R.~Ambrus, N.~Bore, J.~Folkesson,
  P.~Jensfelt, L.~Beyer, A.~Hermans, B.~Leibe, A.~Aldoma, T.~Faulhammer,
  M.~Zillich, M.~Vincze, E.~Chinellato, M.~Al-Omari, P.~Duckworth,
  Y.~Gatsoulis, D.~C. Hogg, A.~G. Cohn, C.~Dondrup, J.~P. Fentanes, T.~Krajnik,
  J.~M. Santos, T.~Duckett, and M.~Hanheide, ``The strands project: Long-term
  autonomy in everyday environments,'' {\em IEEE Robotics Automation Magazine},
  vol.~24, pp.~146--156, Sept 2017.

\bibitem{alterovitz2016robot}
R.~Alterovitz, S.~Koenig, and M.~Likhachev, ``Robot planning in the real world:
  Research challenges and opportunities.,'' {\em AI Magazine}, vol.~37, no.~2,
  pp.~76--84, 2016.

\bibitem{kemp2007challenges}
C.~C. Kemp, A.~Edsinger, and E.~Torres-Jara, ``Challenges for robot
  manipulation in human environments [grand challenges of robotics],'' {\em
  IEEE Robotics Automation Magazine}, vol.~14, pp.~20--29, March 2007.

\bibitem{geman1984gibbs}
S.~Geman and D.~Geman, ``Stochastic relaxation, gibbs distributions, and the
  bayesian restoration of images,'' {\em IEEE Transactions on Pattern Analysis
  and Machine Intelligence}, vol.~PAMI-6, pp.~721--741, Nov 1984.

\bibitem{kucner2013conditional}
T.~Kucner, J.~Saarinen, M.~Magnusson, and A.~J. Lilienthal, ``Conditional
  transition maps: Learning motion patterns in dynamic environments,'' in {\em
  2013 IEEE/RSJ International Conference on Intelligent Robots and Systems},
  pp.~1196--1201, Nov 2013.

\bibitem{wang2014modeling}
Z.~Wang, R.~Ambrus, P.~Jensfelt, and J.~Folkesson, ``Modeling motion patterns
  of dynamic objects by iohmm,'' in {\em 2014 IEEE/RSJ International Conference
  on Intelligent Robots and Systems}, pp.~1832--1838, Sept 2014.

\bibitem{krajnik2017fremen}
T.~Krajník, J.~P. Fentanes, J.~M. Santos, and T.~Duckett, ``Fremen: Frequency
  map enhancement for long-term mobile robot autonomy in changing
  environments,'' {\em IEEE Transactions on Robotics}, vol.~33, pp.~964--977,
  Aug 2017.

\bibitem{endres2013learning}
F.~Endres, J.~Trinkle, and W.~Burgard, ``Learning the dynamics of doors for
  robotic manipulation,'' in {\em 2013 IEEE/RSJ International Conference on
  Intelligent Robots and Systems}, pp.~3543--3549, Nov 2013.

\bibitem{scholz2015learning}
J.~Scholz, M.~Levihn, C.~L. Isbell, H.~Christensen, and M.~Stilman, ``Learning
  non-holonomic object models for mobile manipulation,'' in {\em 2015 IEEE
  International Conference on Robotics and Automation (ICRA)}, pp.~5531--5536,
  May 2015.

\bibitem{wang2002simultaneous}
C.~Wang and C.~Thorpe, ``Simultaneous localization and mapping with detection
  and tracking of moving objects,'' in {\em Proceedings 2002 IEEE International
  Conference on Robotics and Automation (Cat. No.02CH37292)}, vol.~3,
  pp.~2918--2924, 2002.

\bibitem{wang2007simultaneous}
C.~Wang, C.~Thorpe, S.~Thrun, M.~Hebert, and H.~Durrant-Whyte, ``Simultaneous
  localization, mapping and moving object tracking,'' {\em The International
  Journal of Robotics Research}, vol.~26, no.~9, pp.~889--916, 2007.

\bibitem{montemerlo2002conditional}
M.~Montemerlo, S.~Thrun, and W.~Whittaker, ``Conditional particle filters for
  simultaneous mobile robot localization and people-tracking,'' in {\em
  Proceedings 2002 IEEE International Conference on Robotics and Automation
  (Cat. No.02CH37292)}, vol.~1, pp.~695--701 vol.1, 2002.

\bibitem{schulz2001probabilistic}
D.~Schulz and W.~Burgard, ``Probabilistic state estimation of dynamic objects
  with a moving mobile robot,'' {\em Robotics and Autonomous Systems}, vol.~34,
  no.~2, pp.~107--115, 2001.

\bibitem{wolf2003towards}
D.~Wolf and G.~Sukhatme, ``Towards mapping dynamic environments,'' in {\em In
  Proceedings of the International Conference on Advanced Robotics (ICAR)},
  pp.~594--600, 2003.

\bibitem{gallagher2009gatmo}
G.~Gallagher, S.~S. Srinivasa, J.~A. Bagnell, and D.~Ferguson, ``Gatmo: A
  generalized approach to tracking movable objects,'' in {\em 2009 IEEE
  International Conference on Robotics and Automation}, pp.~2043--2048, May
  2009.

\bibitem{mahler2001multitarget}
R.~Mahler and T.~Zajic, ``Multitarget filtering using a multitarget first-order
  moment statistic,'' in {\em Proc. SPIE}, vol.~4380, pp.~184--195, 2001.

\bibitem{vo2006gaussian}
B.~N. Vo and W.~K. Ma, ``The gaussian mixture probability hypothesis density
  filter,'' {\em IEEE Transactions on Signal Processing}, vol.~54,
  pp.~4091--4104, Nov 2006.

\bibitem{pasha2006closed}
A.~Pasha, B.~Vo, H.~d.~Tuan, and W.~k.~Ma, ``Closed form phd filtering for
  linear jump markov models,'' in {\em 2006 9th International Conference on
  Information Fusion}, pp.~1--8, July 2006.

\bibitem{vo2006gaussian2}
B.~N. Vo, A.~Pasha, and H.~D. Tuan, ``A gaussian mixture phd filter for
  nonlinear jump markov models,'' in {\em Proceedings of the 45th IEEE
  Conference on Decision and Control}, pp.~3162--3167, Dec 2006.

\bibitem{punithakumar2008multiple}
K.~Punithakumar, T.~Kirubarajan, and A.~Sinha, ``Multiple-model probability
  hypothesis density filter for tracking maneuvering targets,'' {\em IEEE
  Transactions on Aerospace and Electronic Systems}, vol.~44, pp.~87--98,
  January 2008.

\bibitem{sarkka2007rao}
S.~S{\"a}rkk{\"a}, A.~Vehtari, and J.~Lampinen, ``Rao-blackwellized particle
  filter for multiple target tracking,'' {\em Information Fusion}, vol.~8,
  no.~1, pp.~2--15, 2007.

\bibitem{kreucher2005multitarget}
C.~Kreucher, K.~Kastella, and A.~O. Hero, ``Multitarget tracking using the
  joint multitarget probability density,'' {\em IEEE Transactions on Aerospace
  and Electronic Systems}, vol.~41, pp.~1396--1414, Oct 2005.

\bibitem{dayoub2010toward}
F.~Dayoub, T.~Duckett, G.~Cielniak, {\em et~al.}, ``Toward an object-based
  semantic memory for long-term operation of mobile service robots,'' 2010.

\bibitem{toris2017temporal}
R.~Toris and S.~Chernova, ``Temporal persistence modeling for object search,''
  in {\em 2017 IEEE International Conference on Robotics and Automation
  (ICRA)}, pp.~3215--3222, May 2017.

\bibitem{finman2013toward}
R.~Finman, T.~Whelan, M.~Kaess, and J.~J. Leonard, ``Toward lifelong object
  segmentation from change detection in dense rgb-d maps,'' in {\em 2013
  European Conference on Mobile Robots}, pp.~178--185, Sept 2013.

\bibitem{ambrus2015unsupervised}
R.~Ambrus, J.~Ekekrantz, J.~Folkesson, and P.~Jensfelt, ``Unsupervised learning
  of spatial-temporal models of objects in a long-term autonomy scenario,'' in
  {\em 2015 IEEE/RSJ International Conference on Intelligent Robots and Systems
  (IROS)}, pp.~5678--5685, Sept 2015.

\bibitem{choudhary2014slam}
S.~Choudhary, A.~J.~B. Trevor, H.~I. Christensen, and F.~Dellaert, ``Slam with
  object discovery, modeling and mapping,'' in {\em 2014 IEEE/RSJ International
  Conference on Intelligent Robots and Systems}, pp.~1018--1025, Sept 2014.

\bibitem{collet2015herbdisc}
A.~Collet, B.~Xiong, C.~Gurau, M.~Hebert, and S.~S. Srinivasa, ``Herbdisc:
  Towards lifelong robotic object discovery,'' {\em The International Journal
  of Robotics Research}, vol.~34, no.~1, pp.~3--25, 2015.

\bibitem{sarkka2012backward}
S.~S{\"a}rkk{\"a}, P.~Bunch, and S.~J. Godsill, ``A backward-simulation based
  rao-blackwellized particle smoother for conditionally linear gaussian
  models,'' {\em IFAC Proceedings Volumes}, vol.~45, no.~16, pp.~506--511,
  2012.

\bibitem{shumway1982approach}
R.~H. Shumway and D.~S. Stoffer, ``An approach to time series smoothing and
  forecasting using the em algorithm,'' {\em Journal of time series analysis},
  vol.~3, no.~4, pp.~253--264, 1982.

\bibitem{ekekrantz2017segmentation}
J.~Ekekrantz, N.~Bore, R.~Ambrus, J.~Folkesson, and P.~Jensfelt, ``Unsupervised
  object discovery and segmentation of {RGBD} images,'' {\em arXiv preprint
  arXiv:1710.06929}, 2017.

\bibitem{szegedy2016rethinking}
C.~Szegedy, V.~Vanhoucke, S.~Ioffe, J.~Shlens, and Z.~Wojna, ``Rethinking the
  inception architecture for computer vision,'' in {\em 2016 IEEE Conference on
  Computer Vision and Pattern Recognition (CVPR)}, pp.~2818--2826, June 2016.

\bibitem{maaten2008visualizing}
L.~van~der Maaten and G.~Hinton, ``Visualizing data using t-sne,'' {\em Journal
  of Machine Learning Research}, vol.~9, no.~Nov, pp.~2579--2605, 2008.

\bibitem{bernardin2008evaluating}
K.~Bernardin and R.~Stiefelhagen, ``Evaluating multiple object tracking
  performance: the clear mot metrics,'' {\em EURASIP Journal on Image and Video
  Processing}, vol.~2008, no.~1, pp.~1--10, 2008.

\end{thebibliography}
